\documentclass[preprint,1p,times]{elsarticle}
\usepackage{graphics}
\usepackage{epsfig}
\usepackage{amssymb}

\usepackage{tabularx}
\usepackage{subfig}
\usepackage{xfrac}
\usepackage{multirow}
\usepackage{morefloats}
\usepackage{xspace}

\usepackage{nfontdef}

\newcommand{\ignore}[1]{}

\newcommand{\vek}[1]{\mathchoice{\displaystyle\boldsymbol#1}
{\textstyle\boldsymbol#1}{\scriptstyle\boldsymbol#1}
{\scriptscriptstyle\boldsymbol#1}}

\usepackage{xcolor}

\newcommand{\new}[1]{{\color{blue}#1}}

\def\C{$^{\circ}\mathrm{C}$\xspace}
\newcommand{\ud}{\mathrm{d}}

\journal{Advances in Engineering Software}

\begin{document}

\begin{frontmatter}


\title{Artificial neural networks in calibration
  \\ of nonlinear mechanical models}

\author[ctu]{Tom\'a\v{s} Mare\v{s}}
\ead{marestom87@gmail.com}
\author[ctu]{Eli\v{s}ka Janouchov\'a}
\ead{eliska.janouchova@fsv.cvut.cz}
\author[ctu]{Anna Ku\v{c}erov\'a\corref{auth}}
\ead{anicka@cml.fsv.cvut.cz}
\cortext[auth]{Corresponding author. Tel.:~+420-2-2435-5326;
fax~+420-2-2431-0775}
\address[ctu]{Department of Mechanics, Faculty of Civil Engineering,
  Czech Technical University in Prague, Th\'{a}kurova 7, 166 29 Prague
  6, Czech Republic}
%

\begin{abstract}
Rapid development in numerical modelling of materials and the
complexity of new models increases quickly together with their
computational demands. Despite the growing performance of modern
computers and clusters, calibration of such models from noisy
experimental data remains a nontrivial and often computationally
exhaustive task. The layered neural networks thus represent a robust
and efficient technique to overcome the time-consuming simulations of
a calibrated model. The potential of neural networks consists in
simple implementation and high versatility in approximating nonlinear
relationships. Therefore, there were several approaches proposed to
accelerate the calibration of nonlinear models by neural
networks. This contribution reviews and compares three possible
strategies based on approximating (i) model response, (ii) inverse
relationship between the model response and its parameters and (iii)
error function quantifying how well the model fits the data. The
advantages and drawbacks of particular strategies are demonstrated on
the calibration of four parameters of the affinity hydration model from
simulated data as well as from experimental measurements. This model
is highly nonlinear, but computationally cheap thus allowing its
calibration without any approximation and better quantification of
results obtained by the examined calibration strategies. \new{The
  paper can be thus viewed as a guide intended for the engineers to
  help them select an appropriate strategy in their particular
  calibration problems.}

\end{abstract}


\end{frontmatter}
\section{Introduction}
\label{sec:intro}

Development in numerical modelling provides the possibility to
describe complex phenomena in material or structural behaviour. The
resulting models are, however, often highly nonlinear and defined by
many parameters, which have to be estimated so as to properly describe
the investigated system and its behaviour. The aim of the model
calibration is thus to rediscover unknown parameters knowing the
experimentally obtained response of a system to the given
excitations. The principal difficulty of model calibration is related
to the fact that while the numerical model of an experiment represents
a well-defined mapping from input (model, material, structural, or
other parameters) to output (structural response), there is no
guarantee that the inverse relation even exists.

The most broadly used approach to parameter identification is usually
done by means of an error minimisation technique, where the distance
between parameterised model predictions and observed data is minimised
\cite{Stavroulakis:2003}. Since the inverse relation (mapping of model
outputs to its inputs) is often ill-posed, the error minimisation
technique leads to a difficult optimisation problem, which is highly
nonlinear and multi-modal. Therefore, the choice of an appropriate
identification strategy is not trivial.

Another approach intensively developed during the last decade is based
on Bayesian updating of uncertainty in parameters' description
\cite{Marzouk:2007:JCP,Kucerova:2012:JCAM}. The uncertainty in
observations is expressed by corresponding probability distribution
and employed for estimation of the so-called posterior probabilistic
description of identified parameters together with the prior expert
knowledge about the parameter values
\cite{Jaynes:2003,Tarantola:2005}. The unknown parameters are thus
modelled as random variables originally endowed with prior
expert-based probability density functions which are then updated
using the observations to the posterior density functions. While the
error minimisation techniques lead to a single point estimate of
parameters' values, the result of Bayesian inference is a probability
distribution that summarizes all available information about the
parameters. Another very important advantage of Bayesian inference
consists in treating the inverse problem as a well-posed problem in an
expanded stochastic space.

Despite the progress in uncertainty quantification methods
\cite{Matthies:2007:IB,Rosic:2013:ES}, more information provided by
Bayesian inference is generally related to more time-consuming
computations. In many situations, the single point estimate approach
remains the only feasible one and development of efficient tools
suitable for this strategy is still an actual topic. Within the
several last decades, a lot of attention was paid to the so-called
intelligent methods of information processing and among them
especially to soft computing methods such as artificial neural
networks (ANNs), evolutionary algorithms or fuzzy systems
\cite{Jang:1996:NSC}. A review of soft computing methods for parameter
identification can be found e.g. in \cite{Kucerova:2007:PHD}.  In this
paper, we focus on applications of ANNs in the single point approach
to parameter identification. \new{In particular, we elaborate our
  previous work presented in
  \cite{Mares:2012:IALCCE,Mares:2012:Topping} with the goal to present
  a detail and comprehensive comparison of three different strategies
  of ANNs' usage in parameter identification problems.}

\new{Next section briefly recall the basics of ANNs. Classification of
  ANNs' different applications in calibration problems is introduced
  in Section \ref{sec:strategies} and description of illustrative
  example -- affinity hydration model for concrete -- follows in
  Section \ref{sec:affinity}. In the context of this particular
  example, the calibration strategies are then described in detail in
  five sections starting by training data preparation and sensitivity
  analysis in Section \ref{sec:sensitivity}. Neural network inputs and
  outputs in particular strategies are discussed in Section
  \ref{sec:implementation} and training with topology are described in
  Section \ref{sec:training}. Verification and validation on simulated
  and experimental data are summarized in sections
  \ref{sec:verification} and \ref{sec:valid}, respectively. Finally,
  the results are concluded in Section \ref{sec:concl}.}

\section{Artificial neural network}

Artificial neural networks (ANNs) \cite{Gurney:2002,Haykin:2009} are
powerful computational systems consisting of many simple processing
elements - so-called neurons - connected together to perform tasks
in an analogy to biological brains. Their main feature is the ability to
change their behaviour based on external information that flows
through the ANN during the learning (training) phase.

\begin{figure}[h!]
\centerline{
\includegraphics[width=13cm]{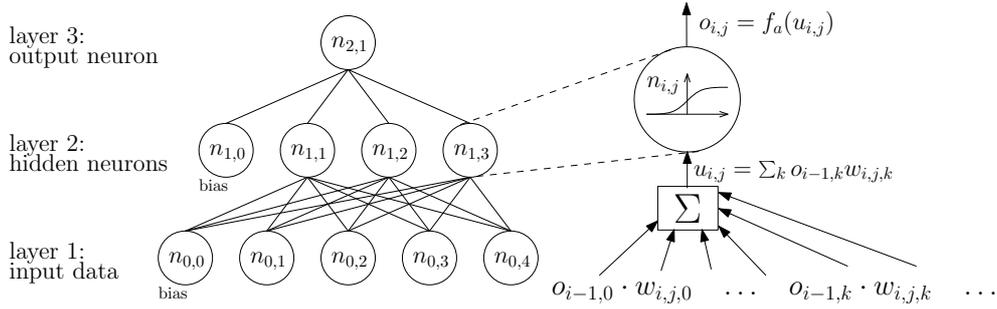} }
\caption{Architecture of multi-layer perceptron}
\label{fig:mlp}
\end{figure}

A particular type of ANN is the so-called feedforward neural network,
which consists of neurons organized into layers where outputs from one
layer are used as inputs into the following layer, see Figure
\ref{fig:mlp}. There are no cycles or loops in the network, no
feed-back connections. Most frequently used example is the multi-layer
perceptron (MLP) with a sigmoid transfer function and a gradient
descent method of training called the back-propagation learning
algorithm.  In practical usage, the MLPs are known for their ability
to approximate nonlinear relations and therefore, when speaking about
an ANN, the MLP is considered in the following text.

The input layer represents a vector of input parameters which are
directly the outputs of the input layer.  The outputs $o_{i-1,k}$ of
the $(i-1)$-th layer are multiplied by a vector of constants
$w_{i,j,k}$, the so-called synaptic weights, summarized and used as
inputs $u_{i,j}$ into \new{the $j$-th neuron in} the following $i$-th layer.
Elements in the hidden and output layers - neurons - are defined by an
activation function $f_a(u_{i,j})$, which is applied on the input and
produces the output value of the $j$-th neuron in the $i$-th layer,
i.e.
\begin{equation}
  o_{i,j} = f_a \left( u_{i,j} \right) \qquad \mbox{where} \qquad u_{i,j} = \new{\sum_{k=0}^K} \left(o_{i-1,k} w_{i,j,k} \right) \, .
\end{equation}
The synaptic weights $w_{i,j,k}$ are parameters of an ANN to be
determined during the training process. \new{$K$ is the number of neurons
in the $i-1$ layer.} The type of the activation function is usually
chosen in accordance with the type of a function to be approximated.
In the case of continuous problems, the sigmoid activation function
given as
\begin{equation}
o_{i,j} = f_a \left( u_{i,j} \right) = \frac{1}{1+e^{-u_{i,j}}}
\end{equation}
is the most common choice.

One bias neuron is usually added into the input and hidden layers. It
does not contain an activation function, but only a constant value.
Its role is to enable to shift the value of a sum over the outputs of
his neighbouring neurons before this sum enters as the input into the
neurons in the following layer. The value of biases is determined by
the training process together with the synaptic weights.

Despite of ANN's popularity there are only few recommendations for the
choice of ANN's architecture. The authors, e.g. in
\cite{Hornik:1989:NN,Hornik:1993:NN}, show that the ANN with any of a
wide variety of continuous nonlinear hidden-layer activation functions
and one hidden layer with an arbitrarily large number of units
suffices for the "universal approximation" property. Therefore, we
limit our numerical experiments to such case.  The number of units in
the input and the output layer is usually given by the studied problem
itself, but there is no theory yet specifying the number of units in
the hidden layer. \new{On one hand, too small number of hidden units
  leads to large prediction errors. On the other hand, a large number of
  hidden units may cause the so-called overfitting, where the ANN
  provides precise outputs for the training samples, but
  fails in case of unseen samples. In such a situation, the ANN tries to
  fit the training data despite increasing oscillations in the
  intermediate space.}

To overcome this problem, some model selection technique
\cite{Anders:1999:NN} has to be applied in order to perform a guided
choice of the ANN's topology. \new{Recent approaches encompass
  e.g. growing-pruning methods (see e.g. \cite{Narasimha:2008:N} or
  more complex techniques designed for optimisation of the ANN's
  topology such as meta-learning
  \cite{Kordik:2009,Kordik:2010:NN}. Here we employ simple and general
  strategy to evaluate a particular ANN's topology: the
  cross-validation}, because it does not involve any probabilistic
assumptions or dependencies on an identification problem.  The idea of
cross-validation is based on a repeated ANN's prediction error evaluation
for a chosen subset of training data and selection of the ANN with the
smallest averaged prediction errors.  Comparing to the well-known
model validation on some independent set of data, the advantage of
cross-validation consists \new{in better behaviour on smaller data sets,
where independent data set cannot be afforded} \cite{Moody:1994:FSNN}.

\new{Before applying the ANN to any engineering problem, one has to
  resolve also several questions regarding the training data
  preparation. It involves not only the transformation of input and
  output data into the range of the activation functions. In
  simulation problems, where the ANN is applied to mimic some unknown
  relationship between observed quantities, the training data coincide
  with the measured data. In inverse problems, we already have some
  theoretical model relating those quantities and we train the ANN on
  simulated data, see a recent review of ANN's application in
  structural mechanics \cite{Freitag:2015:CTR}. Preparation of a
  suitable training set becomes another nontrivial task, where
  sensitivity analysis plays an important role. For a sake of clarity,
  we address these topics in more detail in Section
  \ref{sec:sensitivity} in context with a particular model for
  cement hydration.}

\section{Strategies for application of ANN in model calibration}\label{sec:strategies}

\begin{figure}[h!]
\centerline{
\includegraphics[width=13cm]{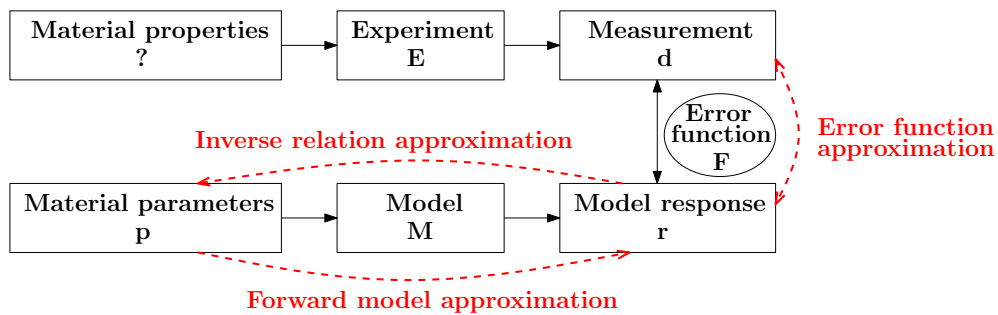} }
\caption{Scheme of model calibration procedure.}
\label{fig:scheme}
\end{figure}

In model calibration, the goal is to find a set of model parameters
minimising the difference between the model response and experimental
measurements, see Figure \ref{fig:scheme}.
An intuitive way of solving the calibration problem is to formulate an
error function quantifying this difference and to minimise the error
function using some optimisation algorithm. The most common error
functions are given as
\begin{eqnarray}
F_1 & = & \sum^{N_\mathrm{R}} _{i=1}(r_i-d_i)^2 \, ,
\label{eq:optim1} \\
F_2 & = & \sum^{N_\mathrm{R}} _{i=1}|r_i-d_i| \, ,
\label{eq:optim2}
\end{eqnarray}
where $r_i$ is the $i$-th component of the model response corresponding to
the $i$-th measured quantity $d_i$ and $N_\mathrm{R}$ is a number of
measured quantities. The difficulty arises from the nonlinear
relation between the model response and model parameters often causing
complexity of the error function such as multi-modality or
non-differentiability. Therefore, the computationally efficient
methods based on analytically or numerically obtained gradient can be
applied only in specific cases.

A more general approach is to apply an optimisation algorithm which
can handle the multi-modality once furnished by a sufficient number of
function evaluations. However, one evaluation of an error function
always involves a simulation of the model. Even for the relatively
fast model simulation, the optimisation can become easily unfeasible
because of the huge number of function evaluations commonly needed by
evolutionary algorithms, even though they usually need less
simulations than uncertainty based methods mentioned in the
introductory part of the paper.

One way of reducing the number of model simulations is to construct a
{\bf forward model approximation} based e.g. on an ANN. The error
function minimisation then becomes a minimisation of distance between
the ANN's predictions and experimental data. The efficiency of this
strategy relies on the evaluation of the trained ANN to be
significantly much faster than the full model simulation.  The
advantage of this strategy is that the ANN is used to approximate a
known mapping which certainly exists and is well-posed.  Computational
costs of this strategy are separated in two parts of a similar size:
(i) the ANN training - optimisation of synaptic weights and (ii) the
minimisation of an error in the ANN prediction for experimental data -
optimisation of ANN inputs (i.e.  determination of investigated model
parameters). An important shortcoming of this method is that this
ill-posed optimisation problem needs to be solved repeatedly for any
new experimental measurement. This way of ANN application to the
parameter identification was presented e.g. in
\cite{Abendroth:2006:EFM}, where an ANN is used for predicting
load-deflection curves and the conjugate directions algorithm is then
applied for optimisation of ductile damage and fracture parameters.
Authors in \cite{Pichler:2003:IJNME} train an ANN to approximate the
results of FE simulations of jet-grouted columns and optimise the
column radius and a cement content of the columns by a genetic
algorithm.  Principally same methods are used for identification of
elasto-plastic parameters in \cite{Aguir:2011:MD}.

One more difficulty of the forward model approximation concerns
the number of parameters and response components. It is very common that
the experimental observations are represented by a discretised curves
or surfaces in time or space dimensions being defined as a vectors
with a large number of components. A forward model then represents a
mapping from usually low-dimensional parameter space to
high-dimensional response space. Although this mapping is well-posed,
the surrogate model must have a large number of outputs or the time
and/or space dimensions have to be included among the model inputs.

Another way of avoiding the mapping to a large number of outputs is to
construct the {\bf error function approximation}, where the model
parameters are mapped onto only one scalar value. One important
inconvenience of such strategy is of course the complexity of the
error function, which can be, as mentioned above, highly nonlinear,
multi-modal and/or non-smooth. Higher complexity of the approximated
relation leads to a higher number of simulations needed for the
construction of the approximation. This concerns another problem of
estimation and choice of an appropriate design of experiments, i.e.
sets of parameters, to perform the simulations which will enable to
build up the surrogate with a relatively small error. This problem can
be reduced by adaptive addition of design points, i.e. new
simulations, close to the minimum of the error function approximation.
The result of the new simulation is then used for an improvement of the
surrogate and a new optimisation process is run again. Such an
approach is usually well suited for surrogates based on kriging or
radial basis function networks \cite{Queipo:2005,Kucerova:2009}. In
this paper, we limit our attention to application of feedforward
layered neural networks and thus, we investigated their ability to
approximate the error function with a limited number of simulations in
non-adaptive fashion.

While the strategy of the forward model approximation involves a new
optimisation process for any new data, the strategy of the error
function approximation involves not only the optimisation process, but
also the surrogate model construction. Regarding this aspect, the most
convenient strategy is the {\bf inverse relation approximation}, which
needs only one evaluation to furnish the parameters corresponding to
new observations. Of course, by the new observations we mean
observations of the system with different properties but performed
under the same external conditions (e.g. a different material, but the
same geometry of the specimen and loading conditions). The strategy of
the inverse relation approximation assumes the existence of an inverse
relationship between the outputs and the inputs of the calibrated
model. If such a relationship exists at least on a specified domain of
parameters' values, it can be approximated by an ANN. Here the ANN's
training process \new{is responsible for all} computational costs
arising from a~solution of the ill-posed problem. This way of the ANN's
application to parameter identification was presented e.g. in
\cite{Novak:2006:EAAN} or recently in~\cite{Kucerova:2014:AES} for
identification of mechanical material parameters, in
\cite{Zaw:2009:JB} for estimation of elastic modulus of the interface
tissue on dental implants surfaces, in \cite{Zhang:2010:ECM} for
identification of interfacial heat transfer coefficient or in
\cite{Klos:2011:CS} for determination of geometrical parameters of
circular arches.

In order to illustrate the advantages and disadvantages of the
outlined strategies of the ANN's application to model calibration, we have
chosen a computationally simple but nonlinear affinity hydration model
briefly described in the following section. The model was successfully
validated on Portland cements in \cite{Silva:2015:JIFS} and thus allows
us to also validate the described identification strategies on
experimental data as summarized in Section \ref{sec:valid}.

\section{Affinity hydration model}\label{sec:affinity}

Affinity hydration models provide a framework for accommodating all
stages of cement hydration. We consider hydrating cement under
isothermal temperature 25\C. At this temperature, the rate of
hydration can be expressed by the {\it chemical affinity}
$\tilde{A}_{25}(\alpha)$ under isothermal 25\C
\begin{equation}
\frac{\ud \alpha}{\ud t}=\tilde{A}_{25}(\alpha),\label{eq:158}
\end{equation}
where the chemical affinity has a dimension of $\textrm{time}^{-1}$
and $\alpha$ stands for the degree of hydration.

The affinity for isothermal temperature can be obtained
experimentally; isothermal calorimetry measures a heat flow $q(t)$
which gives the hydration heat $Q(t)$ after integration. The
approximation is given
\begin{eqnarray}
\frac{Q(t)}{Q_{pot}} &\approx& \alpha\label{eq:146},\\
\frac{1}{Q_{pot}}\frac{\ud Q(t)}{\ud t} &=& \frac{q(t)}{Q_{pot}} \approx \frac{\ud \alpha}{\ud t}= \tilde{A}_{25}(\alpha)\label{eq:45},
\end{eqnarray}
where $Q_{pot}$ is expressed in J/g of cement paste. Hence the
normalized heat flow $\frac{q(t)}{Q_{pot}}$ under isothermal 25\C
equals to chemical affinity $\tilde{A}_{25}(\alpha)$.

Cervera et al. \cite{Cervera:1999:JEM} proposed an analytical form of
the normalized affinity which was refined in
\cite{Gawin:2006:IJNME}. Here we use a slightly modified formulation~\cite{Smilauer:2010}:
\begin{eqnarray}
  \tilde{A}_{25}(\alpha) = B_1 \left( \frac{B_2}{\alpha_\infty} + \alpha \right ) \left( \alpha_\infty - \alpha \right) \exp\left(-\bar\eta\frac{\alpha}{\alpha_\infty}\right),\label{eq:46}
\end{eqnarray}
where $B_1, B_2$ are coefficients related to chemical composition,
$\alpha_\infty$ is the ultimate hydration degree and $\bar\eta$
represents microdiffusion of free water through formed hydrates.

When hydration proceeds under varying temperature, maturity principle
expressed via Arrhenius equation scales the affinity to arbitrary
temperature~$T$
\begin{eqnarray}
  \tilde{A}_{T} = \tilde{A}_{25} \exp\left[\frac{E_a}{R}\left(\frac{1}{273.15+25}-\frac{1}{T}\right)\right],\label{eq:145}
\end{eqnarray}
where $R$ is the universal gas constant (8.314 Jmol$^{-1}$K$^{-1}$)
\new{and $E_a$ [Jmol$^{-1}$] is the activation energy}. For example,
simulating isothermal hydration at 35\C means scaling $\tilde{A}_{25}$
with a factor of 1.651 at a given time. This means that hydrating
concrete for 10 hours at 35\C releases the same amount of heat as
concrete hydrating for 16.51 hours under 25\C. Note that setting
$E_a=0$ ignores the effect of temperature and proceeds the hydration
under 25\C. \new{The evolution of $\alpha$ is obtained through
  numerical integration since there is no analytical exact solution.}

\section{Sensitivity analysis and training data preparation}\label{sec:sensitivity}

Since the ANN's training process requires a preparation of a training data set,
it is also worthy to use these data for a sampling-based sensitivity
analysis \cite{Helton:2006:RESS,Saltelli:2000} and obtain some
information about importance of particular observations or
significance of each parameter for a system behaviour. To achieve some
reliable information from sensitivity analysis as well as a good
approximation by an ANN, one has to choose the training data carefully
according to a suitable design of experiments, see e.g.
\cite{Janouchova:2013:CS} for a competitive comparison of several
experimental designs.

\new{As the model parameters are defined on various intervals, they
  need to be transformed into standardised parameters, e.g. $p_i \in [
    0; 1]$, defined on the intervals suitable for chosen activation
  functions. When the bounds for a parameter vary in orders, it can
  typically suggest highly nonlinear relationship with model
  response. At this moment, any expert knowledge about the parameter
  meaning can be employed to decrease that nonlinearity by
  introduction of nonlinear transformation to standardised parameter.
  It is demonstrated on parameter $B_2$ in Table~\ref{tab:params},
  where bounds for the affinity model parameters together with their
  relations to the standardised parameters $p_i$ are listed.}
\begin{table}[t!]
\centering
\begin{tabular}{lllc}\hline
Parameter & Minimum & Maximum & Relation \\\hline
$B_1 \: [h^{-1}]$      & $0.1 $   & $1$   & $p_1 = (B_1 - 0.1)/0.9$\\
$B_2 \: [-]$           & $10^{-6}$ & $10^{-3}$ & $p_2 = (\log B_2 + 6)/3$\\
$\bar{\eta} \: [-]$    & 2     & 12      & $p_3 = (\bar{\eta} -2)/10$\\
$\alpha_\infty \: [-]$ & 0.7     & 1.0     & $p_4 = (\alpha_\infty -0.7)/0.3$\\
\hline
\end{tabular}
\caption{Bounds for affinity model parameters.}
\label{tab:params}
\end{table}

\begin{figure}[h!]
\centering
\begin{tabular}{cc}
 $B_1 = \{0.1, 0.2, \ldots, 1\}$ & $B_2 = \{10^{-6}, 11.2^{-5}, \ldots, 10^{-3}\}$\\
\includegraphics[width=6cm]{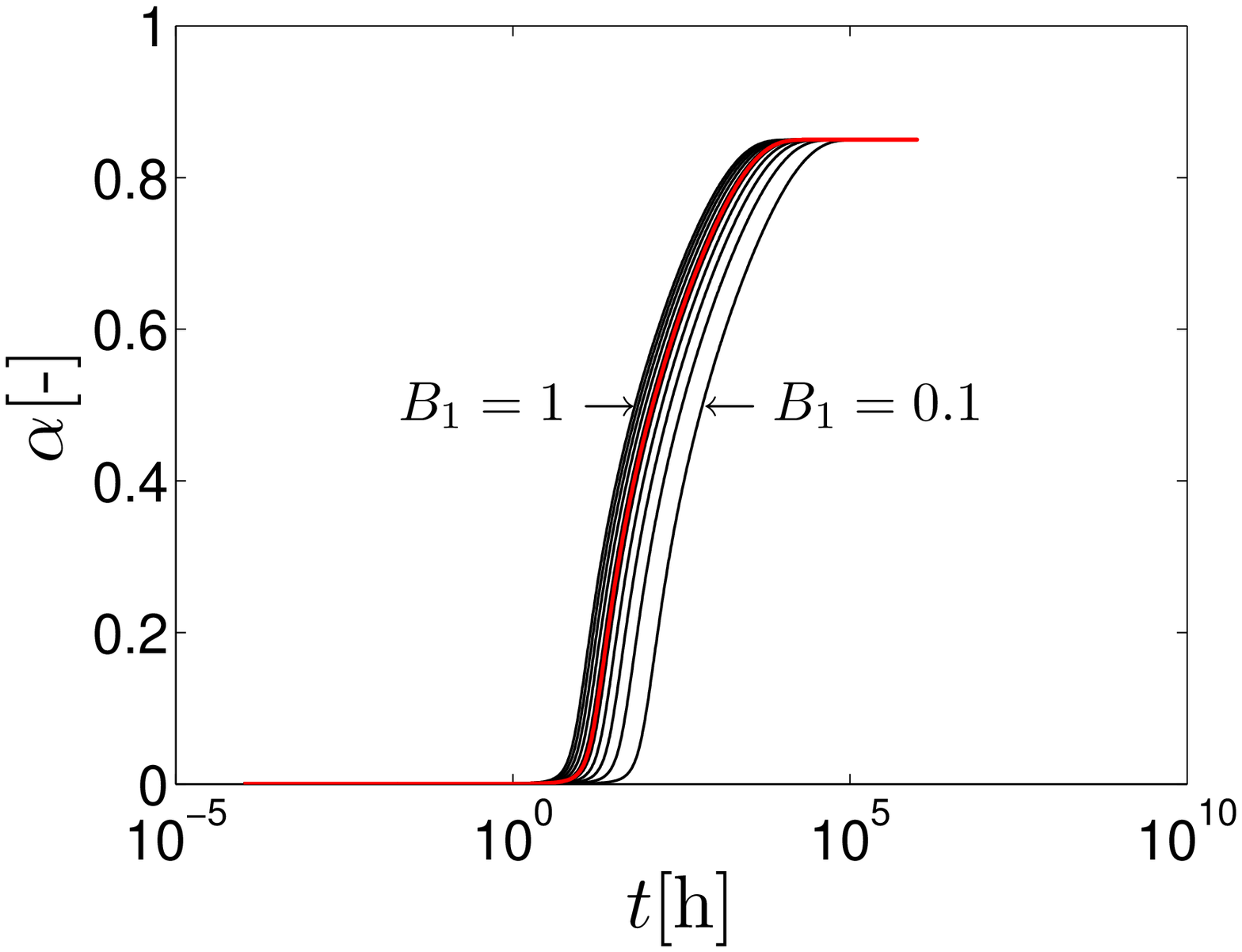} &
\includegraphics[width=6cm]{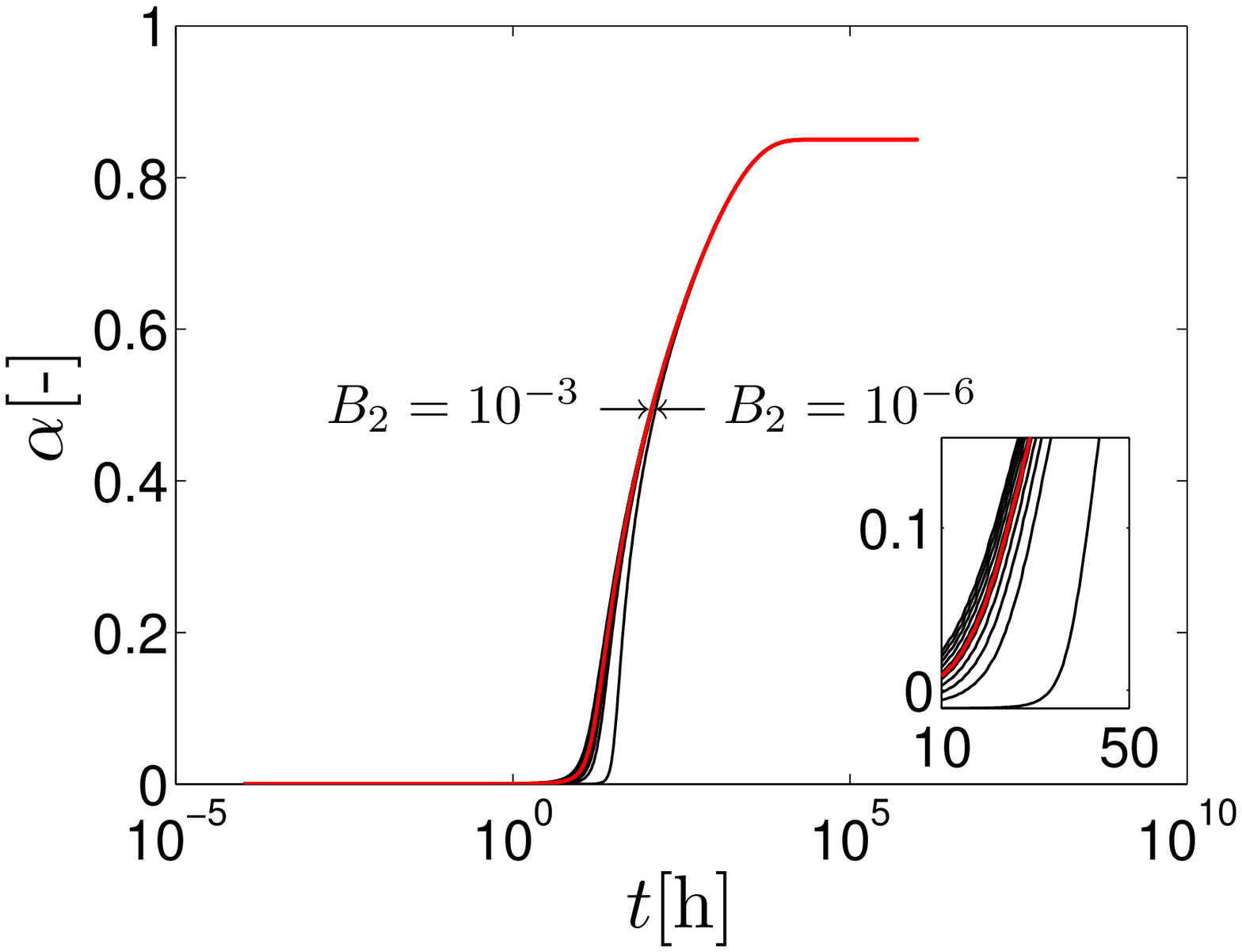}\\
 $\bar{\eta} = \{2, 3.11, \ldots, 12\}$ & $\alpha_\infty = \{0.7, 0.73, \ldots, 1\}$\\
\includegraphics[width=6cm]{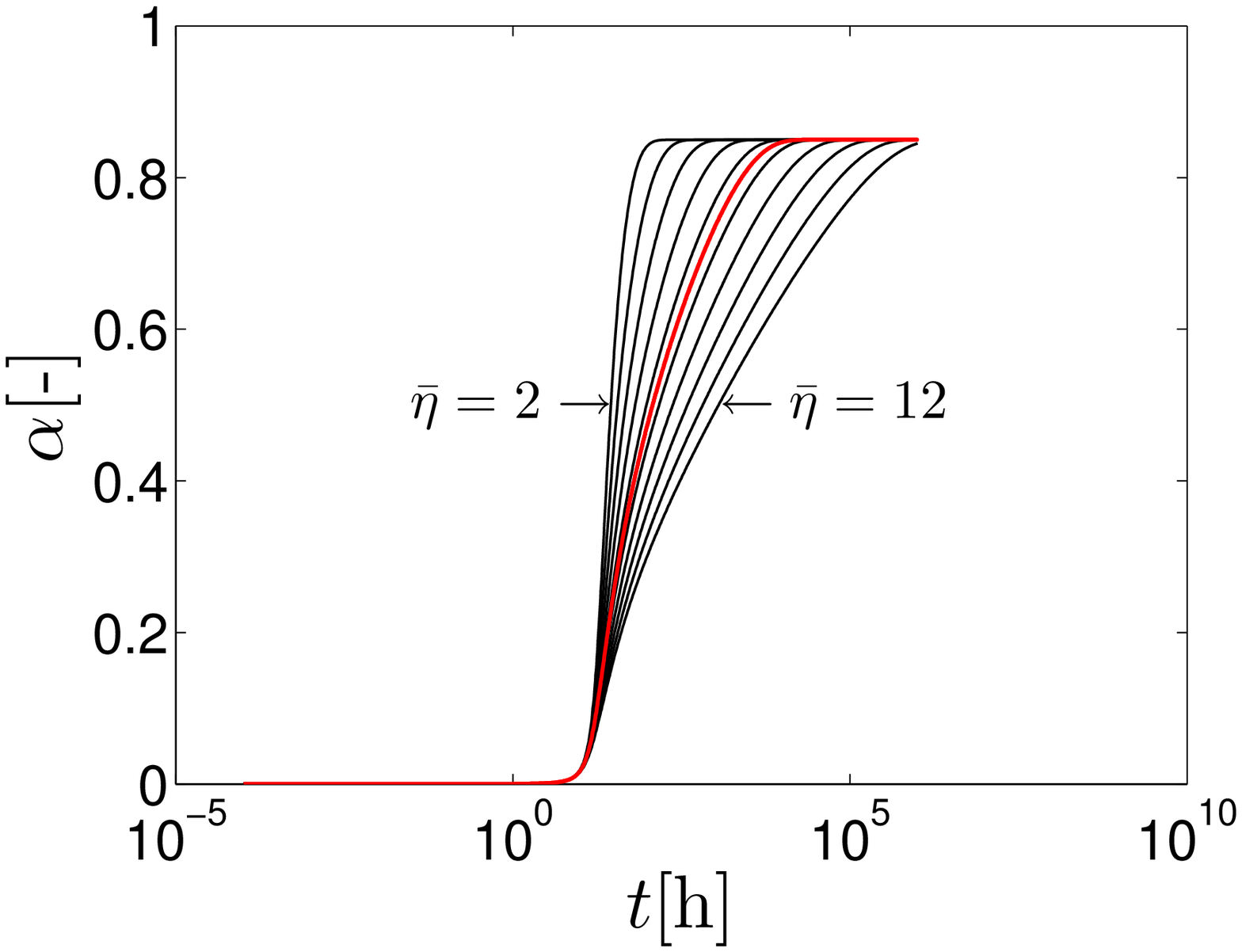} &
\includegraphics[width=6cm]{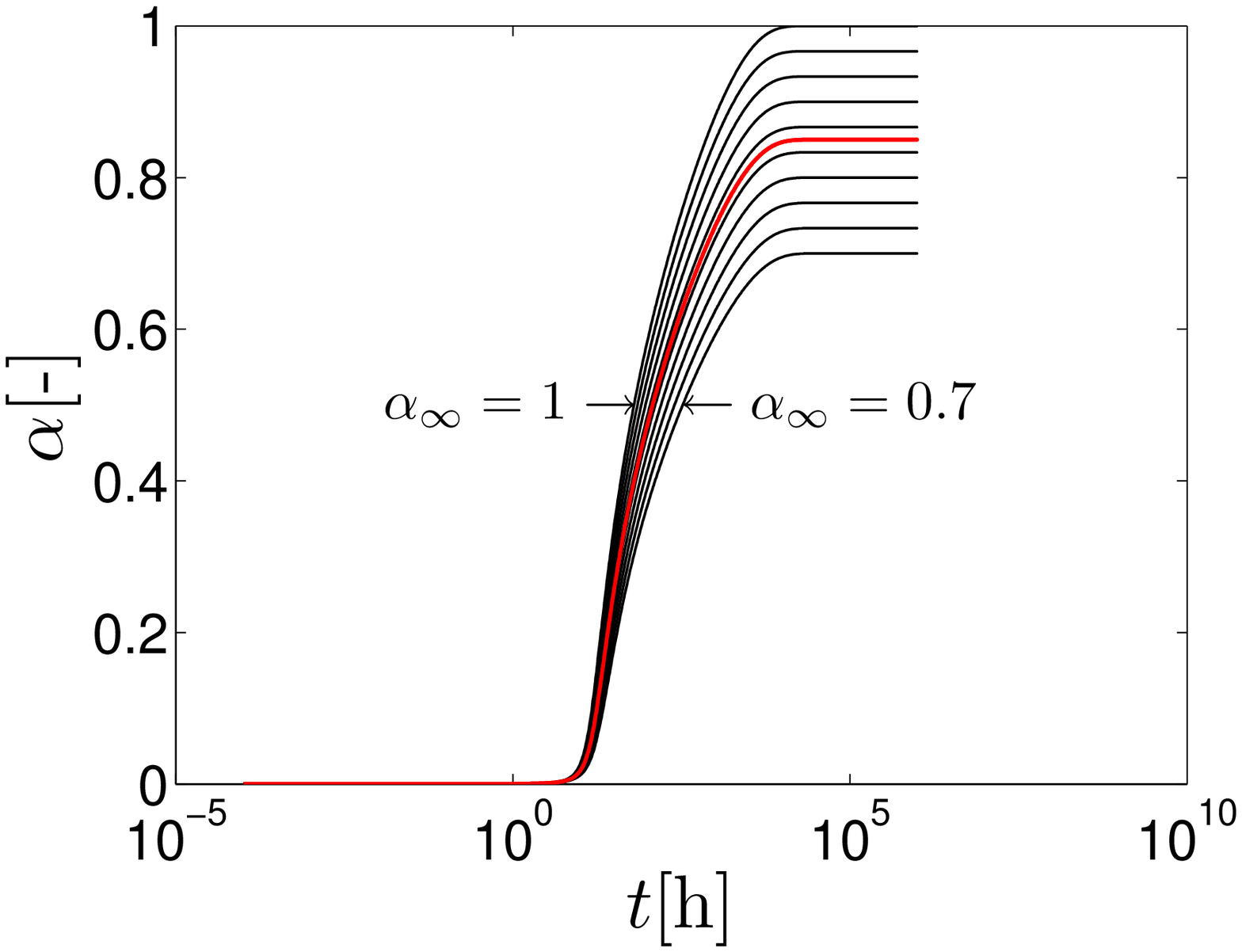}
\end{tabular}
\caption{Influence of model parameters to model response $\alpha$.}
\label{fig:par_alpha}
\end{figure}

\begin{figure}[h!]
\centering
\begin{tabular}{cc}
\includegraphics[width=6cm]{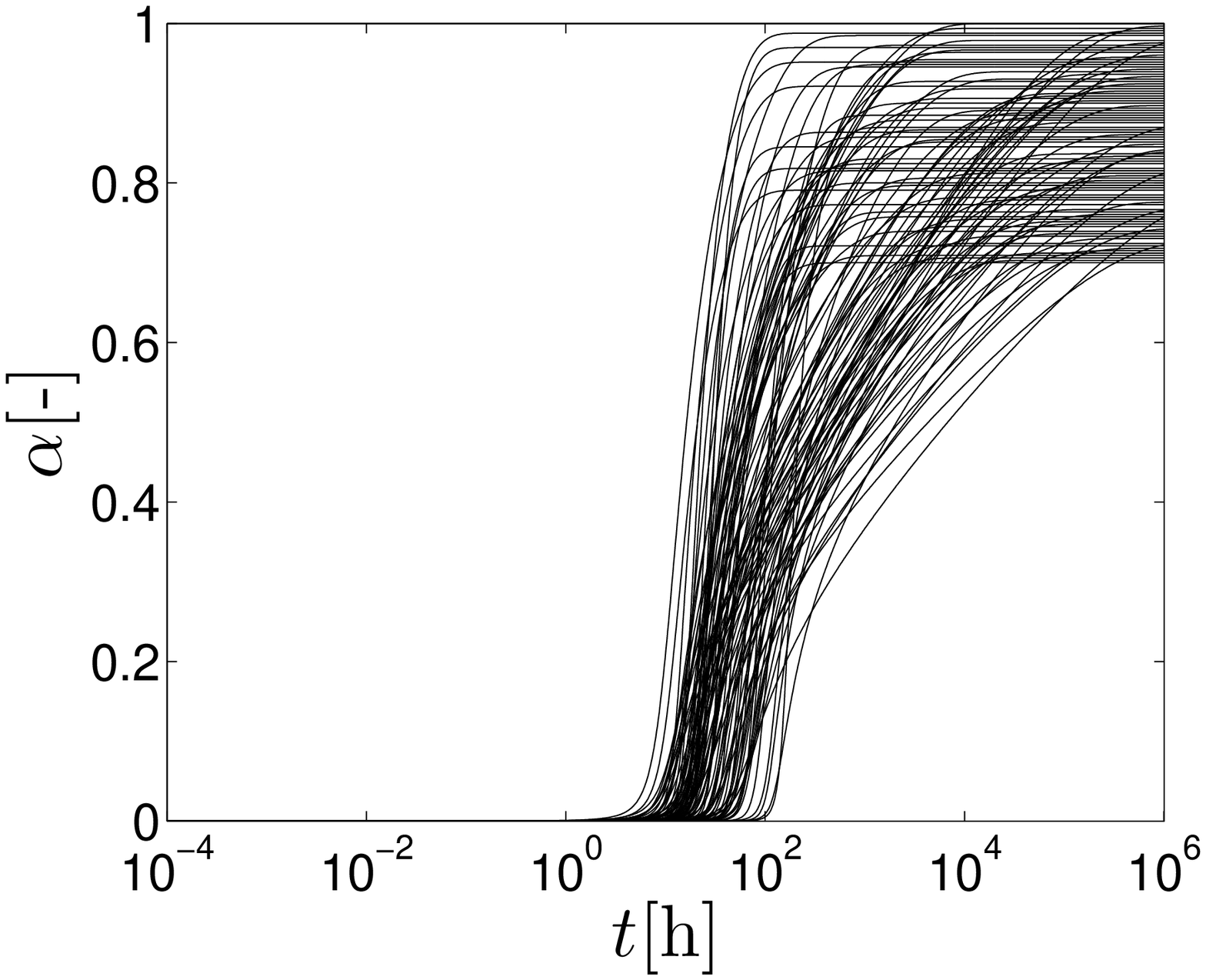} &
\includegraphics[width=6cm]{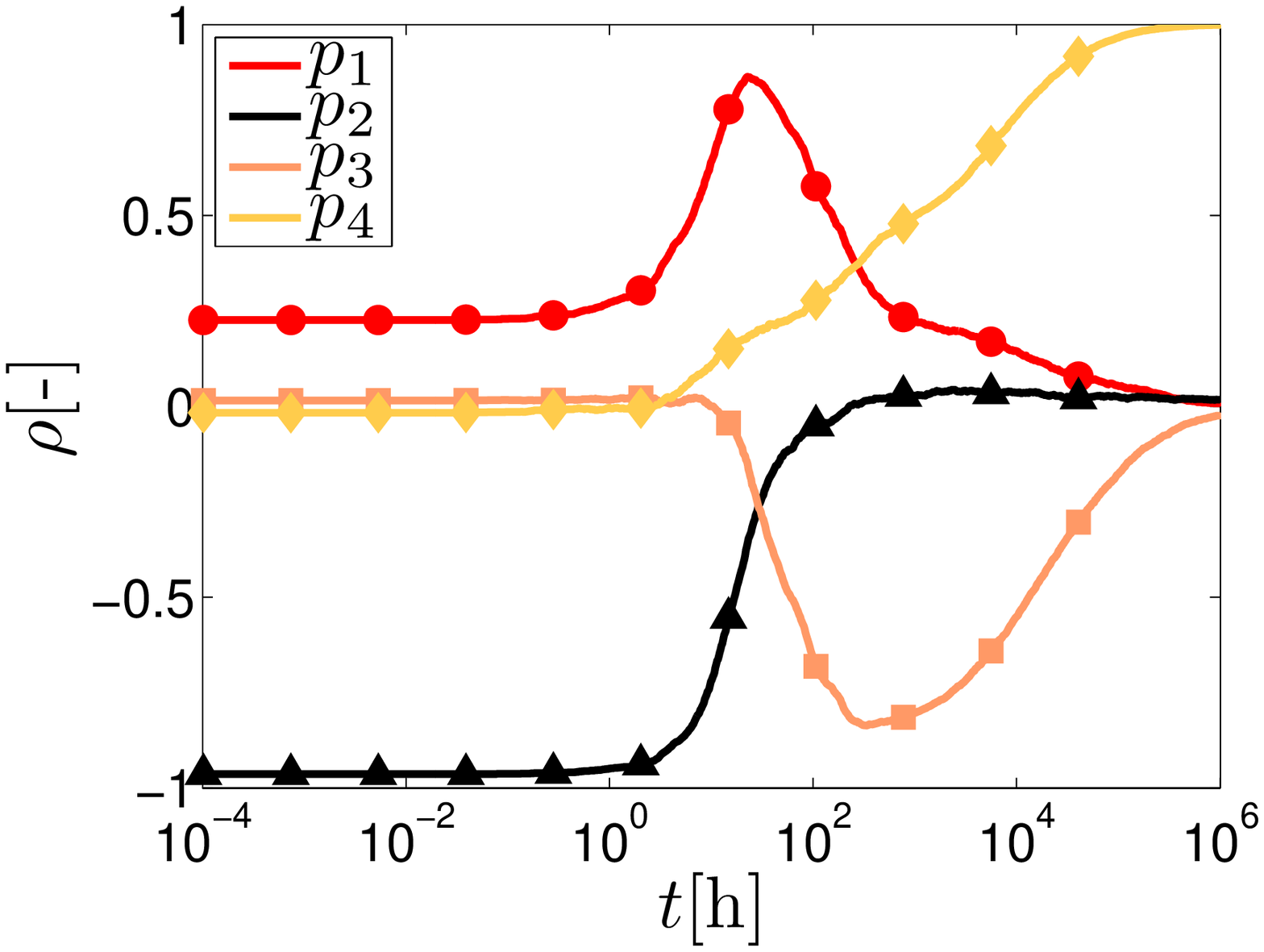} \\
(a) & (b)
\end{tabular}
\caption{Bundle of degree of hydration curves obtained for design points (a) and sensitivity analysis for input-output pairs (b).}
\label{fig:bundle}
\end{figure}

The affinity hydration model was chosen not only for its nonlinearity,
but especially for its relatively simple interpretation and
computationally fast simulation. Hence, we assume that the model is
eligible to illustrate typical features of particular identification
strategies. In order to understand the influence of the model
parameters to its response more deeply, Figure \ref{fig:par_alpha}
demonstrates the changes of the response induced by changes in a
chosen parameter while other parameters are fixed.  On the other hand,
to illustrate the spread of the model response corresponding to the
parameters varying within the given domain, we prepare a design of
experiments (DoE) having $N_\mathrm{DoE} = 100$ samples in the space
of standardised parameters. The DoE is generated as Latin Hypercube
Sampling optimised with respect to the modified $L_2$
discrepancy. Such an experimental design has a~good space-filling
property and is nearly orthogonal~\cite{Janouchova:2013:CS}. For each
design point we perform a~model simulation to obtain a bundle of
$N_\mathrm{DoE}$ curves for the degree of hydration $\alpha(t)$, see
Figure~\ref{fig:bundle}a.

Since the model response is represented by the degree of hydration
being a function of the time, the time domain is discretised into
$1161$ steps uniformly distributed with the logarithm of the time.
Hence, the model input vector $\vek{p} = (p_1, p_2, p_3, p_4)$
consists of $4$ parameters and the output vector $\vek{\alpha} =
(\alpha_1, \dots, \alpha_{N_\mathrm{time}})$ consists of
$N_\mathrm{time} = 1161$ components. In order to quantify the
influence of the model parameters to particular response components, we
evaluate Spearman's rank correlation coefficient $\rho$ for each
$(\vek{p}_i,\vek{\alpha}_i)$ pair using all the $i \in \{1, \dots,
N_\mathrm{DoE}\}$ simulations.  The results of such a~sampling-based
sensitivity analysis \cite{Helton:2006:RESS} are plotted in Figure
\ref{fig:bundle}b.

\begin{figure}[h!]
\centering
\begin{tabular}{cc}
\includegraphics[width=6cm]{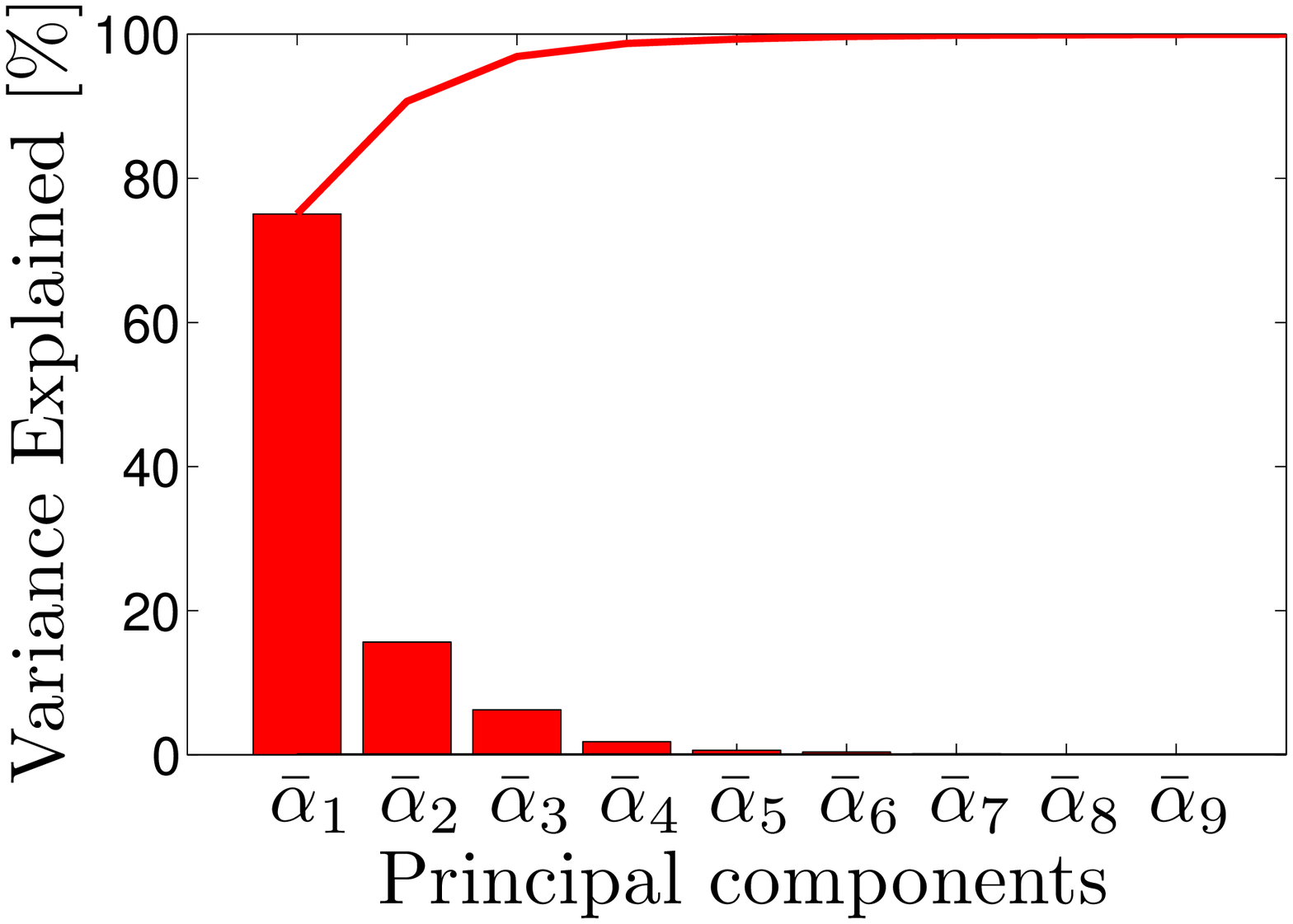} &
\includegraphics[width=6cm]{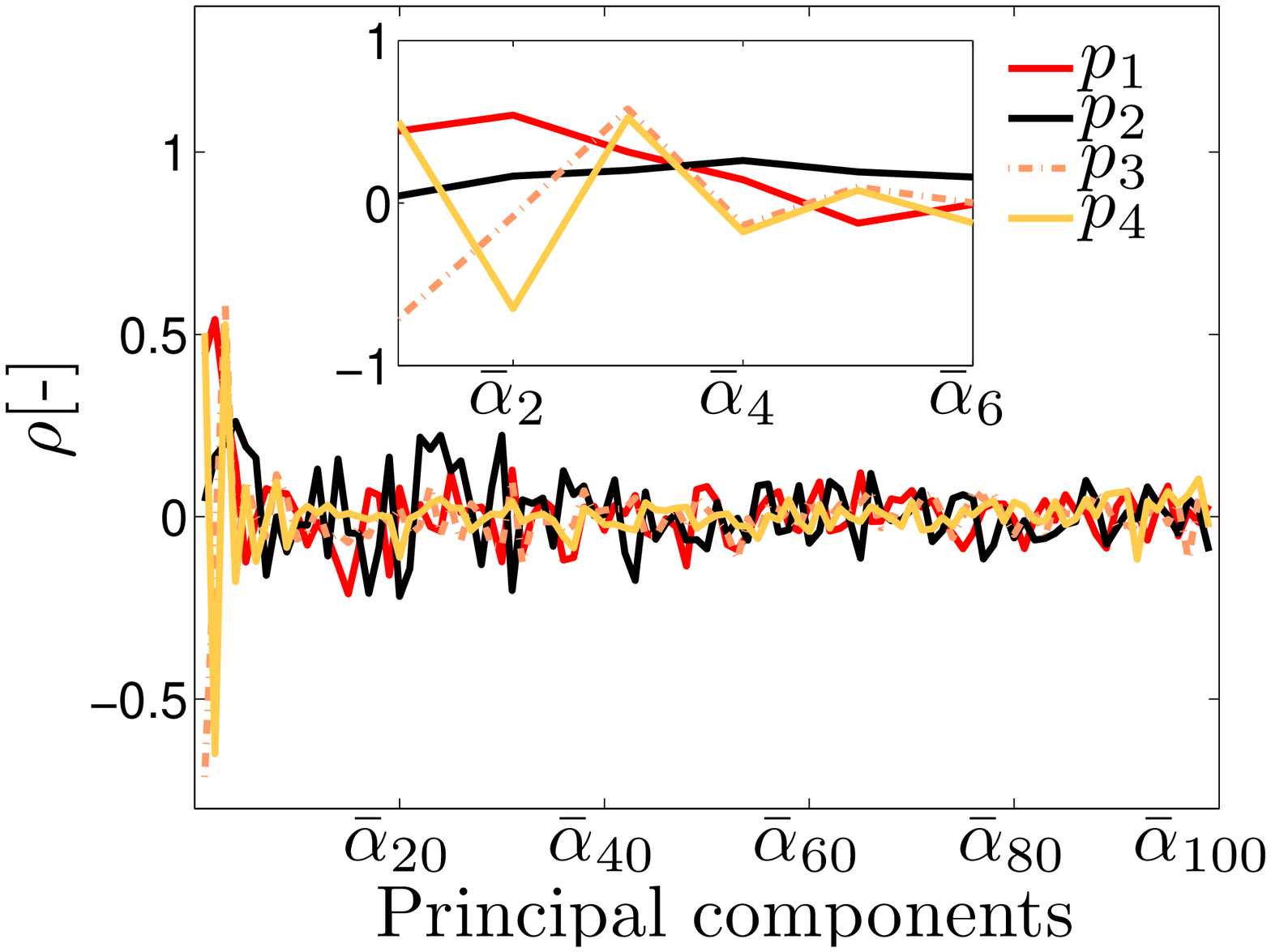} \\
(a) & (b)
\end{tabular}
\caption{Variance explained by the first nine principal components (a)
  and sensitivity analysis for model inputs $\vek{p}_i$ - principal
  components $\bar{\vek{\alpha}}_i$ (b).}
\label{fig:pca}
\end{figure}

In the inverse mode of identification, the model output vector
$\vek{\alpha}$ consisting of $N_\mathrm{time} = 1161$ components is
too large for usage as an input vector for the ANN.  Hence, we
performed the principal component analysis (PCA) in order to reduce
this number to $N_\mathrm{PCA} = 100$ components $\bar{\vek{\alpha}} =
(\bar{\alpha}_1, \dots, \bar{\alpha}_2)$ with non-zero variance (this
number is related to the number of simulations involved in PCA, i.e.
$N_\mathrm{PCA} = N_\mathrm{DoE}$). The components are ordered
according to their relative variance, see Figure \ref{fig:pca}a for
the nine most important ones.
Resulting principal components are technically new quantities obtained
by a linear combination of the original model outputs $\bar{\vek{\alpha}}
= \bar{A}(\vek{\alpha})$. This transformation has of course an
influence to sensitivity analysis and thus we computed correlations
between the model inputs $\vek{p}_i$ and principal components
$\bar{\vek{\alpha}}_i$, see Figure \ref{fig:pca}b.

\section{Implementation of approximation strategies}\label{sec:implementation}

Results of the described simulations are also used as training
simulations for ANNs, i.e. $\mathcal{D}_\mathrm{train} =
\{(\vek{p}_i,\vek{\alpha}_i) \, | \, i \in \{1, 2, \dots,
N_\mathrm{train}\}, N_\mathrm{train} = N_\mathrm{DoE} = 100 \}$.
Particular approximation strategies, however, process the training
simulations in a different way.

The strategy of the forward model approximation can be formulated in two
ways, which differ in handling the high dimensionality of the model output
$\vek{\alpha}$.  In the first formulation, we can consider the
time step $t_k$ as the fifth model parameter (i.e. the fifth model input)
and thus the model output reduces into only one scalar value of
the hydration degree $\alpha_k$ corresponding to the given time $t_k$. As
the objective of the ANN is thus to span the parameter as well as the
time space, we called this strategy as {\bf Forward Complex}
(ForwComp). In such a configuration, the results of $N_\mathrm{train}$
training simulations turn into $N_\mathrm{train} \times
N_\mathrm{time} = 116,100$ training samples.  Evaluation of so many
samples at every iteration of the ANN's training process is, however, very
time-consuming.  Therefore, only every  $m$-th time step
is included for ANN training and thus the training set is given as
$\mathcal{D}_\mathrm{train}^\mathrm{ForwComp} =
\{((\vek{p}_i,t_k),\alpha_{i,k}) \, | \, i \in \{1, 2, \dots,
N_\mathrm{train}\}, k \in \{1, 1+m, 1+2m, \dots, N_\mathrm{time}\}
\}$.  In our particular implementation, we selected $m=10$ leading to
$|\mathcal{D}_\mathrm{train}^\mathrm{ForwComp}| = 11,700$ samples.
Note that in all other strategies, the number of training samples
equals the number of training simulations, see Table \ref{tab:stratpar},
where the significant parameters of particular approximation
strategies are briefly summarised.

\begin{table}[h!]
\centering
\begin{tabular*}{\textwidth}{@{\extracolsep{\fill} }lrllr}
  \hline
  Strategy & $N_\mathrm{ANN}$ & Inputs & Outputs & $|\mathcal{D}_\mathrm{train}|$ \\
  \hline
  Forward Complex   & $1$ & $p_1$, $p_2$, $p_3$, $p_4$, $t_k$ & $\alpha_k \, | \, k \in \{1, 11, \dots, 1161\}$ & $11700$ \\
  Forward Split     & $9$  & $p_1$, $p_2$, $p_3$, $p_4$ & $\alpha_{300}; \alpha_{400}; \dots; \alpha_{1100}$ & $100$ \\
  Forward Split II  & $22$ & $p_1$, $p_2$, $p_3$, $p_4$ & $\alpha_{100}; \alpha_{150}; \dots; \alpha_{1150}$ & $100$ \\
  Forward Split III & $43$ & $p_1$, $p_2$, $p_3$, $p_4$ & $\alpha_{100}; \alpha_{130}; \alpha_{150}; \alpha_{170}; \dots; \alpha_{1150}$ & $100$ \\
\hline
  Error $F_1$       & $1$ & $p_1$, $p_2$, $p_3$, $p_4$ & $F_1$ & $100$ \\
  Error $F_2$       & $1$ & $p_1$, $p_2$, $p_3$, $p_4$ & $F_2$ & $100$ \\
\hline
  Inverse Expert    & $4$ & $\alpha_{300}, \alpha_{400}, \dots, \alpha_{1100}$ & $p_1$; $p_2$; $p_3$; $p_4$ & $100$ \\
  Inverse Expert II & $4$ & $\alpha_{200}, \alpha_{300}, \dots, \alpha_{1100}$ & $p_1$; $p_2$; $p_3$; $p_4$ & $100$ \\
  Inverse PCA       & $4$ & $\bar{\alpha}_1, \bar{\alpha}_2, \dots, \bar{\alpha}_9$ & $p_1$; $p_2$; $p_3$; $p_4$ & $100$ \\
  \hline
\end{tabular*}
\caption{Parameters of approximation strategies}
\label{tab:stratpar}
\end{table}

The second way of the model output approximation is based on training an
independent ANN for every time step $t_k$. Here, the particular ANN
approximates simpler relation and span only the parameter space. A
training data set for ANN approximating the response component
$\alpha_k$ is thus given as
$\mathcal{D}_\mathrm{train}^{\mathrm{ForwSpli,}\alpha_k} =
\{(\vek{p}_i,\alpha_{i,k}) \, | \, i \in \{1, 2, \dots, 100\} \}$
having only 
$|\mathcal{D}_\mathrm{train}^{\mathrm{ForwSpli,}\alpha_k}| = 100$
samples.  A disadvantage of such an approach consists in training
a large number $N_\mathrm{ANN}$ of smaller ANNs. As training of
$N_\mathrm{ANN} = N_\mathrm{time} = 1161$ different ANNs can be almost
unfeasible, we select only a few of the time steps, where the
approximation is constructed and thus, the model output approximation
is more rough.  The choice of the important time steps and their
number can be driven by the expert knowledge or results of the
sensitivity analysis.  Hence, we present three different choices so as
to illustrate its influence, see Table \ref{tab:stratpar}. We further
call these strategies as {\bf Forward Split} (ForwSpli), {\bf Forward
  Split II} (ForwSpliII) and {\bf Forward Split III} (ForwSpliIII).

The error function approximation is the only strategy where the high
dimensionality of the model output does not impose any complications. The
model output is used for evaluation of the error function
and the ANN is trained to approximate the mapping from the parameter
space to a single scalar value of the error function, i.e.
$\mathcal{D}_\mathrm{train}^{\mathrm{Error,}F_a} = \{(\vek{p}_i,F_a)
\, | \, i \in \{1, 2, \dots, N_\mathrm{train}\}\}$ and
$|\mathcal{D}_\mathrm{train}^{\mathrm{Error,}F_a}| = 100$, where $F_a$
stands for a chosen error function. As we already mentioned in Section
\ref{sec:strategies}, there are two very common error functions given
by Eqs. \eqref{eq:optim1} and \eqref{eq:optim2} and thus we
investigate both considering the two strategies further called as {\bf
  Error $F_1$} and {\bf Error $F_2$}, respectively.

In case of the inverse relation approximation, the high dimensionality of
the model output needs again some special treatment so as to keep the
number of ANN inputs and thus the ANN complexity reasonable. An
intuitive approach is a simple selection of a limited number of output
values $\vek{a} = A(\vek{\alpha})$. Here, one ANN is trained to
predict one model parameter $p_j$ and thus
$\mathcal{D}_\mathrm{train}^{\mathrm{InvExp},p_j} =
\{(\vek{a}_i,p_{i,j}) \, | \, i \in \{1, 2, \dots,
N_\mathrm{train}\}\}$ and
$|\mathcal{D}_\mathrm{train}^{\mathrm{InvExp,}p_j}| = 100$. A
particular choice of components in the vector $\vek{a}_i$ defined by
the operator $A$ should take into account not only the results of
sensitivity analysis, but also a possible measurement error in
experimental data as well as any other expert knowledge.
Hence we present again two different choices in order to illustrate
its influence, see Table \ref{tab:stratpar} and we further call
these configurations as {\bf Inverse Expert} (InvExp) and {\bf Inverse
  Expert II} (InvExpII).

In order to reduce the influence of the expert choice, the principal
components $\bar{\vek{\alpha}}$ computed as described in the previous
section can be used as the ANN's inputs and one has to choose only their
number. To compare the information contained in the same number of
inputs selected by an expert, we have chosen the same number of
principal components as the number of inputs in the Inverse Expert
configuration and thus
$\mathcal{D}_\mathrm{train}^{\mathrm{InvPCA},p_j} =
\{((\bar{\alpha}_{i,1}, \dots, \bar{\alpha}_{i,9}),p_{i,j}) \, | \, i
\in \{1, 2, \dots, N_\mathrm{train}\}\}$ and
$|\mathcal{D}_\mathrm{train}^{\mathrm{InvPCA,}p_j}| = 100$.  The
principal components based strategy is further called {\bf Inverse
  PCA} (InvPCA). \new{In our preliminary study presented in
  \cite{Mares:2012:IALCCE}, we have also tested the possibility to
  choose smaller number of PCA-based inputs selected separately for
  each parameter to be identified according to the sensitivity
  analysis. Nevertheless, such sensitivity-driven reduction of
  PCA-based inputs was shown to deteriorate the quality of trained
  ANNs.}

Then, the last preparatory step concerns the generation of testing
data for a final assessment of the resulting ANNs consisting of
$N_\mathrm{test} = 50$ simulations for randomly generated sets of
input parameters. The obtained data are then processed by particular
approximation strategies in the same way as the training data
described above.

\section{Neural network training algorithm and topology choice}\label{sec:training}

The quality of the ANN-based approximation estimated on a given data set
$\mathcal{D}$ can be expressed as the mean relative prediction error
$\varepsilon^\mathrm{MRP}(\mathcal{D})$ given as
\begin{equation}
  \varepsilon^\mathrm{MRP}(\mathcal{D}) = \frac{\sum^{|\mathcal{D}|}_{i=1}|O_{i}-T_{i,\mathcal{D}}|}{|\mathcal{D}|( T_\mathrm{max,\mathcal{D}_{train}}-T_\mathrm{min,\mathcal{D}_{train}})} \, ,
\label{eq:error}
\end{equation}
where $O_{i}$ is the ANN's output corresponding to the target value
$T_{i,\mathcal{D}}$ contained in the data set $\mathcal{D}$, which
consists of $|\mathcal{D}|$ samples.
$T_\mathrm{max,\mathcal{D}_{train}}$ and
$T_\mathrm{min,\mathcal{D}_{train}}$ are the maximal and minimal
target values in the training data set $\mathcal{D}_\mathrm{train}$, so the
error $\varepsilon^\mathrm{MRP}(\mathcal{D})$ is always scaled by the
same factor for any chosen data set $\mathcal{D}$ and this factor
corresponds to the range of the training data.

The conjugate gradient-based method \cite{Shewchuk:1994} was applied
as a training algorithm for synaptic weights computation and the
cross-validation method was employed to determine the number of hidden
neurons. In $V$-fold cross-validation we break the training data set
$\mathcal{D}_\mathrm{train}$ into $V$ approximately equisized subsets
$\mathcal{D}_\mathrm{train} = \mathcal{D}_{\mathrm{train},1} \cup
\mathcal{D}_{\mathrm{train},2} \cup \dots \cup
\mathcal{D}_{\mathrm{train,}V}$ and then we perform $V$ training
processes, each time leaving out one of the subsets
$\mathcal{D}_{\mathrm{train},i}$ and using the rest of the training
data set $\mathcal{D}_\mathrm{train} \setminus
\mathcal{D}_{\mathrm{train,i}}$.

The criterion for stopping the training process is governed by the
prediction errors ratio $r^\mathrm{PE}_k$ computed at the $k$-th
iteration of the training algorithm given as
\begin{equation}
  r^\mathrm{PE}_k (\mathcal{D}_\mathrm{train} \setminus
  \mathcal{D}_{\mathrm{train,i}}) = \frac{ \sum_{j = k-J}^{k} \varepsilon^\mathrm{MRP}_j (\mathcal{D}_\mathrm{train} \setminus
    \mathcal{D}_{\mathrm{train,i}}) }{ \sum_{j = k-2J}^{k-J-1} \varepsilon^\mathrm{MRP}_j (\mathcal{D}_\mathrm{train} \setminus
    \mathcal{D}_{\mathrm{train,i}})} \, ,
\end{equation}
where $\varepsilon^\mathrm{MRP}_j (\mathcal{D}_\mathrm{train}
\setminus \mathcal{D}_{\mathrm{train,i}})$ is the mean relative
prediction error obtained at the $j$-th iteration of the training
algorithm obtained on the training data set without its $i$-th partition.
$J$ is the chosen number of iterations considered for computing the
ratio $r^\mathrm{PE}_k$ for its smoothing effect on
$r^\mathrm{PE}_k$. The training process is stopped either when the
number of iterations achieves its chosen maximal value $K$ or if the
prediction errors ratio $r^\mathrm{PE}_k$ exceeds a chosen critical
value $r^\mathrm{PE}_\mathrm{max}$.

Once the training process is completed, the ANN is evaluated on the
remaining part of the training data $\mathcal{D}_{\mathrm{train,i}}$,
which was not used in the training process. The quality of the ANN with a
particular number of hidden neurons $h$ is assessed by the
cross-validation error $\varepsilon^\mathrm{CV}_h$, which is computed
as a mean of the errors obtained for the ANNs trained on the subsets
$\mathcal{D}_\mathrm{train} \setminus \mathcal{D}_{\mathrm{train,i}}$
and then evaluated on the remaining subset
$\mathcal{D}_{\mathrm{train,i}}$ , i.e.
\begin{equation}
  \varepsilon^\mathrm{CV}_h = \frac{1}{V} \sum_{i=1}^{V} \varepsilon^\mathrm{MRP} (\mathcal{D}_{\mathrm{train,i}}) \, .
\end{equation}
We start with an ANN having $h_\mathrm{min}$ hidden neurons and we
compute the corresponding cross-validation error. Then, one hidden
neuron is added and after all the training processes on training data
subsets, the new cross-validation error is evaluated. We compute the
cross-validation error ratio $r^\mathrm{CVE}_h$ as
\begin{equation}
  r^\mathrm{CVE}_h = \varepsilon^\mathrm{CV}_h /
  \varepsilon^\mathrm{CV}_{h-1} \, .
\end{equation}
We count the situations when the ratio $r^\mathrm{CVE}_h$ exceeds a
chosen critical value $r^\mathrm{CVE}_\mathrm{max}$. If this happened
$W$ times, the addition of hidden neurons is stopped. Then we choose
the architecture having the smallest cross-validation error
$\varepsilon^\mathrm{CV}_h$ and the particular ANN with the synaptic
weights having the smallest training error $\varepsilon^\mathrm{MRP}$.

\begin{table}[h!]
\centering
\begin{tabular*}{\textwidth}{@{\extracolsep{\fill} }lcc}
  \hline
  Number of subsets in cross-validation & $V$ & $10$ \\
  Number of iteration considered in $r^\mathrm{PE}_k$ & $J$ & 100 \\
  Maximal number of training iterations & $K$ & $5000$ \\
  Maximal value of prediction errors ratio & $r^\mathrm{PE}_\mathrm{max}$ & $0.999$ \\
  Starting value of hidden neurons & $h_\mathrm{min}$ & $1$ \\
  Maximal value of cross-validation error ratio & $r^\mathrm{CVE}_\mathrm{max}$ & $0.99$ \\
  Maximal value of $r^\mathrm{CVE}_\mathrm{max}$ exceeding & $W$ & $3$ \\
  \hline
\end{tabular*}
\caption{Parameters of ANN training algorithm and cross-validation method}
\end{table}

The resulting ANNs are tested on an independent testing data set
$\mathcal{D}_\mathrm{test}$. Since some of the approximation
strategies consist of a high number of ANNs, the resulting number of
hidden neurons and achieved errors on training and testing data for
all the trained ANNs are listed in \ref{app:ann_config}.  Brief
summary of these results is presented in Table
\ref{tab:training}\footnote{The error function approximation
  strategies are intrinsically related to particular experimental curve.
  The results here are obtained for experimental "Mokra" data described
  in Section \ref{sec:valid} in more detail.}.
\begin{table}[h!] \tabcolsep=2pt
\centering
\begin{tabular*}{\textwidth}{@{\extracolsep{\fill}}lrrr}\hline
  Strategy & $h$ & $\varepsilon^\mathrm{MRP}(\mathcal{D}_\mathrm{train}) [\%]$ & $\varepsilon^\mathrm{MRP}(\mathcal{D}_\mathrm{test}) [\%]$\\
  \hline
  Forward Complex   & $7$                        & $2.03$                          & $2.67$                         \\
  Forward Split     & $3$ to $10$                & $0.06$ to \hspace{0.9mm} $1.06$ & $0.06$ to \hspace{0.9mm} $1.27$\\
  Forward Split II  & $4$ to $13$                & $0.06$ to \hspace{0.9mm} $1.42$ & $0.07$ to \hspace{0.9mm} $2.04$\\
  Forward Split III & $3$ to $13$                & $0.03$ to \hspace{0.9mm} $1.50$ & $0.03$ to \hspace{0.9mm} $1.98$\\
\hline
  Error $F_1$       & $10$                       & $0.40$ to \hspace{0.9mm} $0.54$ & $0.57$ to \hspace{0.9mm} $0.74$\\
  Error $F_2$       & $9$ to $11$                & $0.78$ to \hspace{0.9mm} $1.36$ & $0.96$ to \hspace{0.9mm} $1.56$\\
\hline
  Inverse Expert    & $5$ to \hspace{0.9mm} $8$  & $1.14$ to \hspace{0.9mm} $5.74$ & $1.31$ to \hspace{0.9mm} $6.43$\\
  Inverse Expert II & $4$ to \hspace{0.9mm} $6$  & $1.38$ to \hspace{0.9mm} $5.79$ & $1.36$ to \hspace{0.9mm} $6.52$\\
  Inverse PCA       & $4$ to \hspace{0.9mm} $8$  & $0.28$ to               $10.50$ & $0.33$ to               $16.73$\\
  \hline
\end{tabular*}
\caption{Architecture of particular ANNs in inverse strategies and their errors on training and testing data.}
\label{tab:training}
\end{table}

Regarding the number of hidden neurons, the results point to higher
complexity of the error function relationships. Nevertheless, the
differences in hidden neurons among particular strategies are
relatively small.

The quality of the resulting ANNs in approximation of the given relationships
is measured by the obtained errors on all the training
$\varepsilon^\mathrm{MRP}(\mathcal{D}_\mathrm{train})$ and testing
$\varepsilon^\mathrm{MRP}(\mathcal{D}_\mathrm{test})$ data. Small
differences between the training and testing errors refer to
well-trained ANNs and to the good quality of the training method as
well as the method for topology estimation. Note that overtrained ANNs
usually lead to significantly higher errors on testing data.

Comparing the approximation quality of the particular strategies, we can
point out good results of the forward model approximation and error
function approximation, where the errors did not exceed the value of $3$
\%. The good approximation of the forward model is not surprising since the
relationship is well-defined, smooth and relatively simple. The good
results of the error function approximation are more unexpected, because
the relationship here is probably more nonlinear and complex. One
possible explanation is a large spread of error function values on
the training data, which is used to scale the errors (see Eq.
\eqref{eq:error}). While the error functions converge to zero near the
optimal parameter values, they quickly rise to extremely high values
for parameter values more distant from the optimum. Hence, we presume
that the small errors obtained in the error function approximation do not
promise comparably good results in the final parameter identification.

The results of the inverse relation approximation are not very good, but
it was foreseen due to unknown and probably ill-posed relationship.
Nevertheless, the obtained errors are actually the final errors of the
whole identification process for the training and testing data, since there
is no other following step concerning any optimisation as in the case
of other identification strategies. Hence, further comments on these
results are presented in the following section concerning verification
of the overall identification strategies on the testing data.

\section{Verification of model calibration}\label{sec:verification}

Since the errors in Table \ref{tab:training} represent only the
quality of the constructed ANNs, we have to also investigate the quality
of the identification procedures. This section is devoted to verification
of the model calibration, where the goal is to predict the model
parameters' values corresponding to the simulated data, which are not
perturbated by any noise. The advantage of verification is that we
also know the true values of the parameters and thus, we can easily
evaluate the quality of their estimation by each strategy. In
particular, the calibration strategies were applied to estimate the
parameters' values for all the training and testing simulations.

As mentioned, in case of the inverse relation approximation, the outputs
of ANNs are directly the predicted values of the identified parameters
$\widehat{\vek{p}}$.
In case of the forward model approximation, we have to run a
subsequent optimisation process. Here, the evolutionary algorithm
GRADE, see \cite{Kucerova:2007:PHD} for details about this
method\footnote{The parameters of GRADE algorithm were set to
  pool\_rate = 4, radioactivity = 0.33 and cross\_limit = 0.1. The
  algorithm was stopped after $10000$ cost function evaluations.}, is
applied to find a set of parameters' values $\widehat{\vek{p}}$
minimising the square distance $\delta$ between components of the model
response $\alpha_k$ and their corresponding ANN-based approximated
counterparts $\widetilde{\alpha}_k$, i.e.
\begin{equation}
\delta = \sum_k (\alpha_k - \widetilde{\alpha}_k)^2 \, ,
\label{eq:dist}
\end{equation}
where $k$ corresponds to the selected approximated components defined for
particular identification strategies in Table \ref{tab:stratpar}.  In
such a way, the parameters $\widehat{\vek{p}}$ are predicted for all
the training as well as testing data. As the true values of parameters
$\vek{p}$ are known in the verification process, the mean prediction
errors $\widehat{\varepsilon}$ are computed relatively to the spread
of the training data, i.e.
\begin{equation}
\widehat{\epsilon}(\widehat{p}_j) = \frac{\sum_{i=1}^{|\mathcal{D}|}|p_{i,j} -
  \widehat{p}_{i,j}|}{|\mathcal{D}|(p_{\mathrm{max}(\mathcal{D}_{\mathrm{train}}),j} -
  p_{\mathrm{min}(\mathcal{D}_{\mathrm{train}}),j})} \, ,
\label{eq:prederr}
\end{equation}
and the obtained errors for particular identification strategies are
listed in Table \ref{tab:ident}.
\begin{table}[h!]
\begin{tabular}{l|rr|rr|rr|rr|rr}
\hline
& \multicolumn{2}{l|}{$\widehat{\varepsilon}(\widehat{p}_1)$} & \multicolumn{2}{l|}{$\widehat{\varepsilon}(\widehat{p}_2)$} & \multicolumn{2}{l|}{$\widehat{\varepsilon}(\widehat{p}_3)$} & \multicolumn{2}{l|}{$\widehat{\varepsilon}(\widehat{p}_4)$} & \multicolumn{2}{l}{$\widehat{\varepsilon}(\widehat{\alpha})$}\\
& train & test & train & test & train & test & train & test & train & test\\\hline
Forward Complex      & 16.78 & 17.09  & 52.20 & 47.91    & 6.06 & 5.45    & 3.67 & 2.69    & 1.079 & 1.088  \\
Forward Split      & 9.48 & 11.62   & 30.18 & 38.45    & 3.14 & 4.65    & 1.17 & 3.10    & 0.310 & 0.370 \\
Forward Split II    & 5.09 & 6.47    & 13.34 & 15.03    & 1.69 & 2.60    & 0.67 & 1.02    & 0.144 & 0.205 \\
Forward Split III    & 4.12 & {\bf 4.84}    & 10.73 & 10.65    & 1.49 & {\bf 1.63}    & 0.57 & 0.64    & {\bf 0.124} & {\bf 0.160} \\
Inverse Expert        & 5.74 & 6.43    & {\bf 5.15} & {\bf 6.21}      & 1.99 & 2.16    & 1.14 & 1.31    & 0.490 & 0.493  \\
Inverse Expert II      & 5.79 & 6.23    & 5.60 & 6.52      & 2.60 & 3.18    & 1.38 & 1.36    & 0.444 & 0.533  \\
Inverse PCA        & {\bf 3.86} & 5.10    & 10.50 & 16.73    & {\bf 1.25} & 1.89    & {\bf 0.28} & {\bf 0.33}    & 0.377 & 1.209  \\
\hline
\end{tabular}
\caption{Results of verification of particular identification
  strategies in terms of mean relative prediction errors
  $\widehat{\varepsilon}$ [\%]. \new{Best results are highlighted in bold font.}}
\label{tab:ident}
\end{table}

In application of the identification strategy to real experimental data,
the parameter values are not known, but the success of the
identification process is quantified by quality of fitting the data by
the model response obtained for the identified parameters. Hence, the model
simulations were performed for all the identified parameter sets and
prediction errors $\widetilde{\varepsilon}$ in terms of predicted
responses $\widetilde{\vek{\alpha}}$ are computed analogously to the
Eq.~\eqref{eq:prederr}. Their values averaged also over all the
response components are then listed in Table~\ref{tab:ident}.

The results for the strategies based on an approximation of the error function
are missing here, because they require to build a particular ANN for
every curve of the hydration degree and for each require to run an
additional minimisation procedure. This is overwhelming and thus these
strategies are only validated on the experimental data as described in the following section.

One can see that among the forward strategies, the complex variant
provided the worst results in the training process as well as in the final
identification. The complex relationship covering the time domain
causes apparently certain difficulties to the training process. We can
conclude that training of a set of neural networks means more work,
but offers significantly better quality of the model approximation. We
can also point out the large differences in errors of particular
parameters, which correspond to influence of particular parameters to
the model response. As demonstrated in Figure~\ref{fig:par_alpha}, the
largest spread of the model response is related namely to change in
the parameters $p_4$ and $p_3$, while the parameter $p_1$ and also $p_2$
seem to be almost negligible. The sensitivity analysis illustrated in
Figure~\ref{fig:bundle}b shows very high sensitivity of the model response
to the parameter $p_2$ at early stage of hydration, nevertheless, at this
stage the spread of the model response is almost negligible and even a very
small error in the response approximation can be fatal for identification
of the parameter $p_2$. On the other hand, it is not surprising that the
identification accuracy is significantly improved with an increasing
number of approximated response components, i.e. an increasing number of
trained ANNs.

Despite the worse results in training of ANNs, the inverse strategies
achieved comparably good results with the forward strategies in parameter
identification and also in fitted measurements. More precisely, the
results of measurements fitting are slightly worse, but the errors in
parameter prediction are smaller. Especially the Inverse Expert
strategies provided surprisingly small errors in $p_2$ prediction and
the errors in parameters are generally more balanced. This phenomenon
can be possibly explained by fact that each ANN is trained to predict
each parameter separately, thus automatically selecting and
emphasizing the combinations of the model response critical for the
parameter. In the strategy Inverse Expert II, the usage of one
additional input at the early stage of hydration caused no improvement
of the resulting prediction, which is probably caused again by fact
that the responses at this stage have a negligible spread and almost no
predictive value. The last interesting result concerns the application of
principal component analysis. The Inverse PCA strategy provided again
significantly different errors in prediction of particular parameters,
similarly to the forward strategies. The reason resides possibly in fact
that PCA emphasize the most important components, while it can mix the
effects of the less significant parameters. Nevertheless, when
compared with strategies Forward Split and Inverse Expert using the
same number of response components, the Inverse PCA provided the best
results in prediction of all the parameters except $p_2$. Its quality
of measurement fitting is, however, the worst among those
strategies.

From this thorough comparison we may conclude that all the inverse
strategies provide very good results, which makes them highly
promising considering their very simple implementation which does not
include any additional optimisation process except the only training
of ANNs. Moreover, the Inverse Expert strategies can be especially
recommended for identification of less significant parameters.

\section{Validation of model calibration}\label{sec:valid}

The previous section was focused on mutual comparison of the presented
identification strategies on simulated data. However, a complete
comparison has to include their validation on experimental data.  To
that purpose we used the four experimental data obtained by isothermal
calorimetry: one for cement ``{\bf Mokra}'' CEM I 42.5 R taken
directly from Heidelberg cement group's kiln in Mokr\'a, Czech
Republic \cite{Smilauer:2010} and three others from the following
literature: ``{\bf Boumiz}'' \cite{Boumiz}, ``{\bf Hua}'' \cite{Hua}
and ``{\bf Princigallo}'' \cite{Princigallo}.

\begin{figure}[h!]
\begin{tabular}{cc}
\includegraphics[width=7cm]{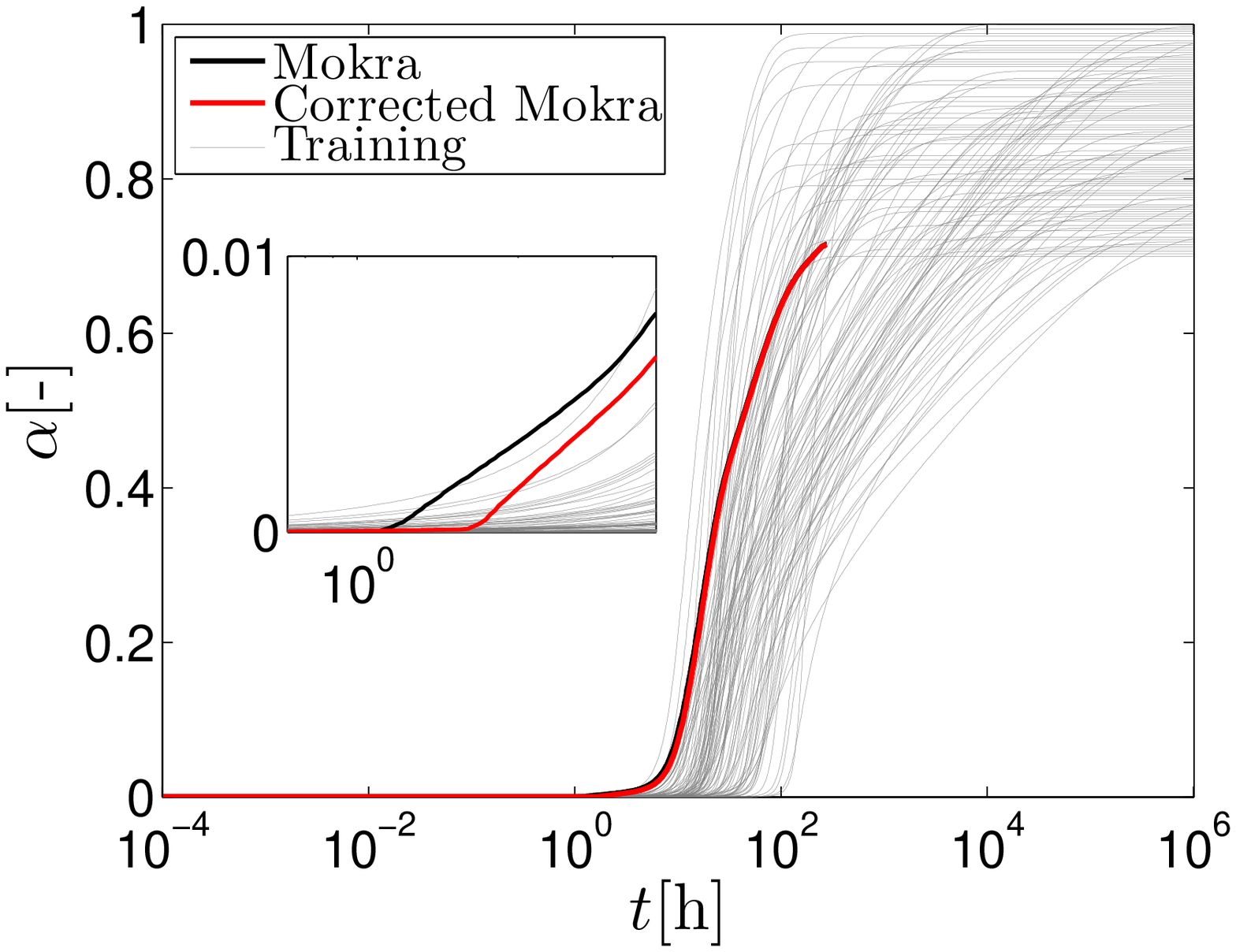} &
\includegraphics[width=7cm]{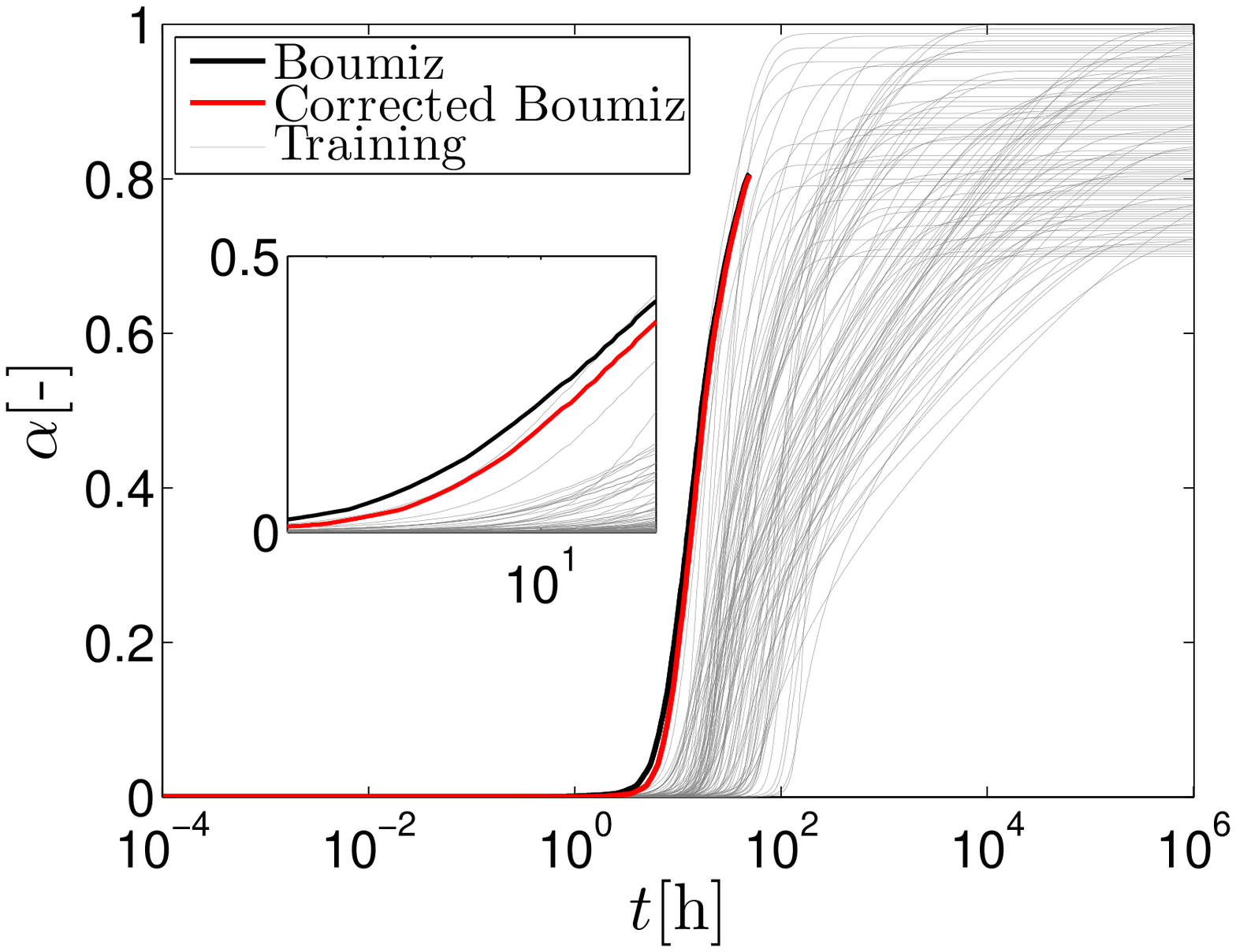} \\
correction: $0.5$ h & correction: $1$ h \\
\includegraphics[width=7cm]{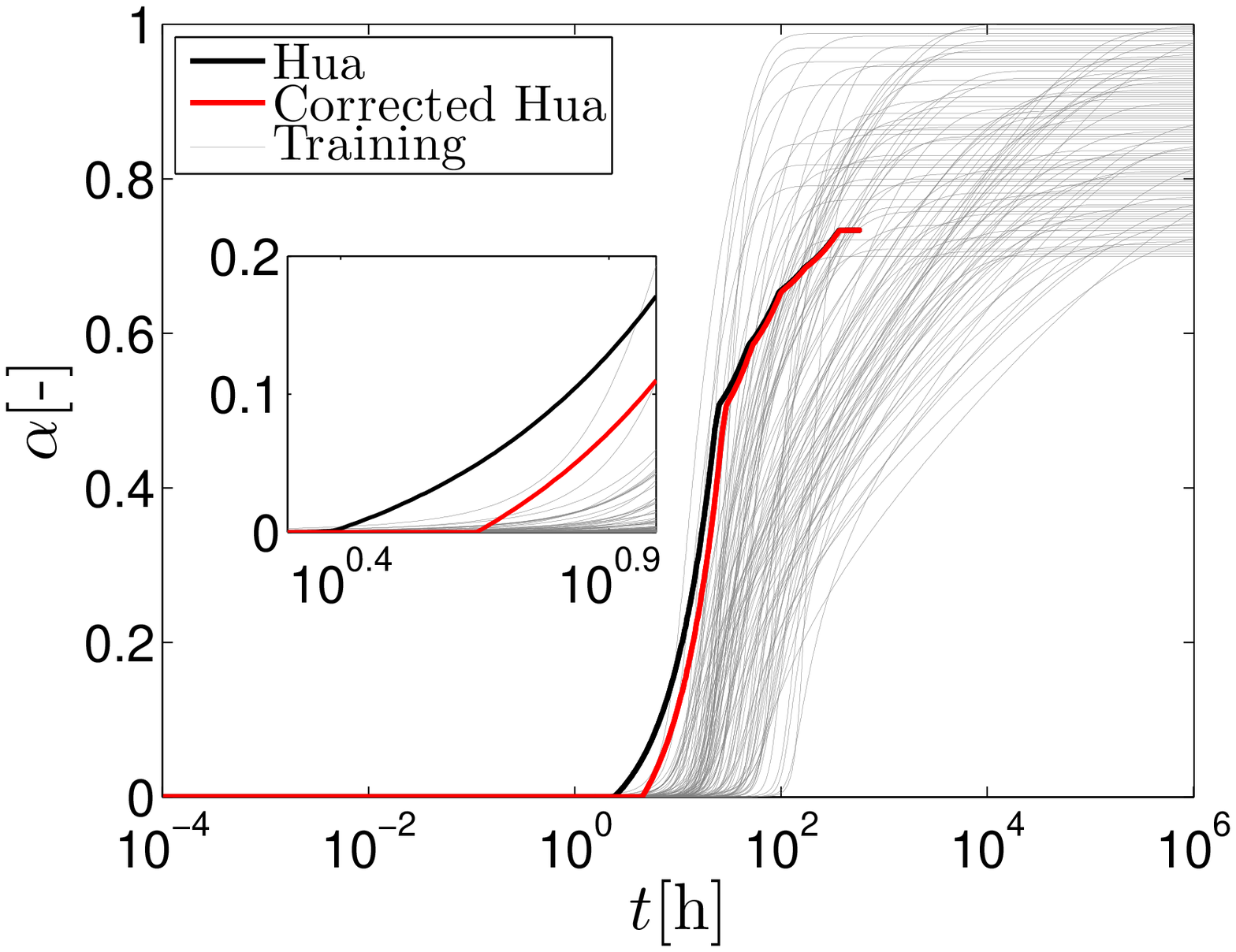} &
\includegraphics[width=7cm]{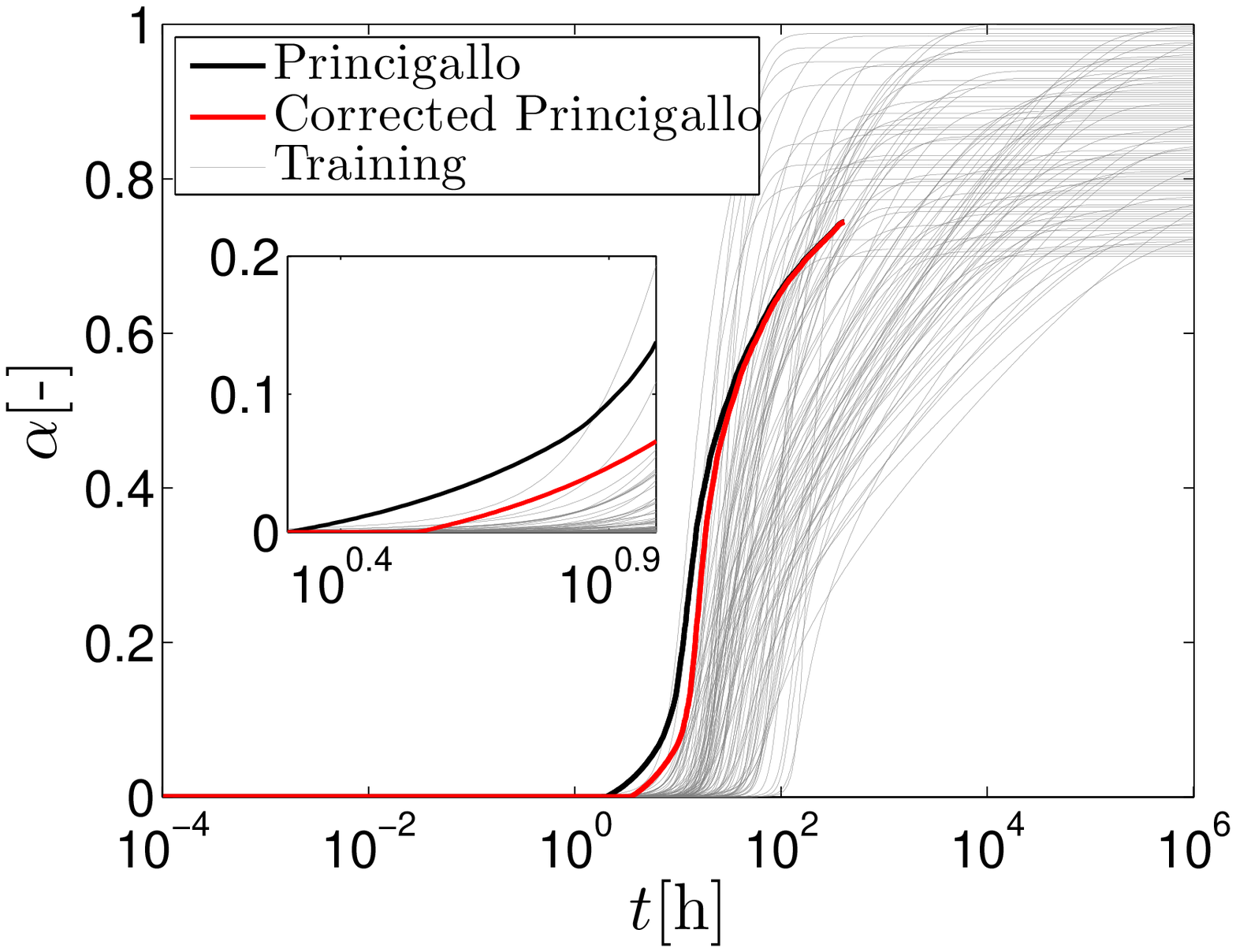}\\
correction: $4.5$ h & correction: $3.5$ h
\end{tabular}
\caption{Corrections of experimental curves.}
\label{fig:shifts}
\end{figure}

In parameter identification from experimental data, one often face to
difficulties related to (i) experimental errors and (ii) model
imperfections. Especially in case of models with parameters having a
specific physical meaning -- like the affinity hydration model -- it
happens that the experimental data seems to lie beyond the physically
meaningful values of the model parameters. This is exactly what we
face in case of the four experimental curves depicted in
Figure~\ref{fig:shifts}.
The grey curves represent the training samples generated in an
optimised fashion so as to maximally cover the parameter
space. Nevertheless, it is visible that all the experimental curves
lie out of the bundle of the training samples. Applying the
identification strategies to these data will require the ANNs to
extrapolate and it will probably lead to unphysical and wrong
predictions of the model parameters. Such results were presented for
``Mokra'' in \cite{Mares:2012:Topping}. Looking in more detail on the
experimental curves, one can see that the difference between the
experimental data and simulations can be explained by wrong estimation
of the origin of hydration. Correction of the starting time moves the
curves into the bundle of response simulations.  As a matter of fact,
the correction in orders of hours is negligible comparing to the
duration of the whole hydration process lasting often days or
weeks. Moreover, the goal of this paper is not to argue against the
correctness of the model or data, but to demonstrate the properties of
particular identification strategies which can be better illustrated
in a situation, where the observed data are not outliers w.r.t. sampled
parameter domain. For an interested reader about the identification of
outliers we refer to \cite{Mares:2012:Topping}.

In general, validation does not allow for a comparison in terms of
parameters' values, because these are not known a priori.
Nevertheless, the simplicity and the fast simulation of the affinity
hydration model permit a direct optimisation of the model parameters
so as to fit the measured data without any incorporated approximation.
The resulting optimal solutions can be then compared with the results
obtained using the ANN approximations. To that purpose, we employ
again the error functions given in Eqs.~\eqref{eq:optim1} and
\eqref{eq:optim2} and the GRADE algorithm with the same setting as in
the previous section to minimise the both error functions. The
obtained results are referred to as {\bf Direct1} and {\bf Direct2},
respectively, \new{and they represent the best results that can be achieved with the current model on the given data.}

\begin{figure}[h!]
\begin{tabular}{cc}
\includegraphics[width=7cm]{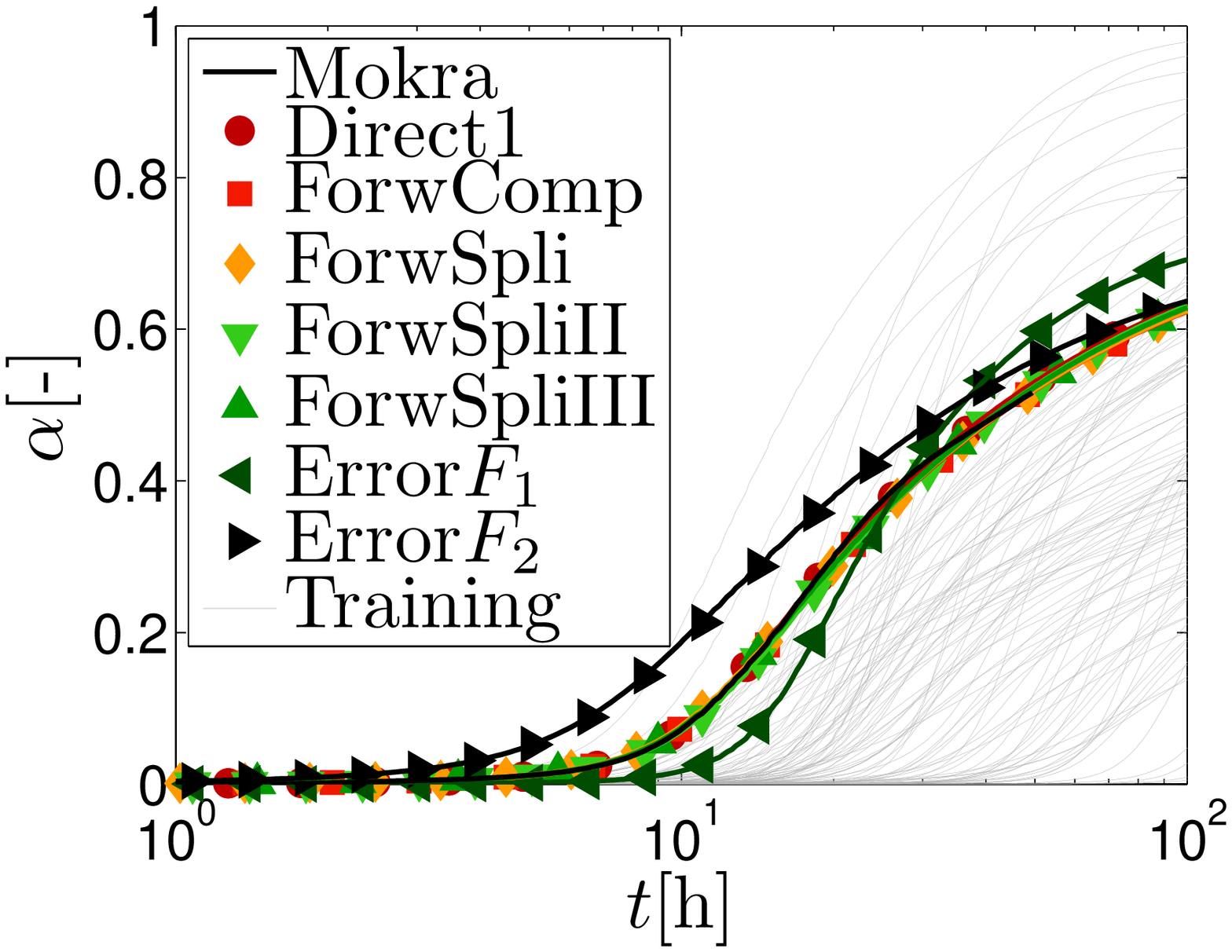} &
\includegraphics[width=7cm]{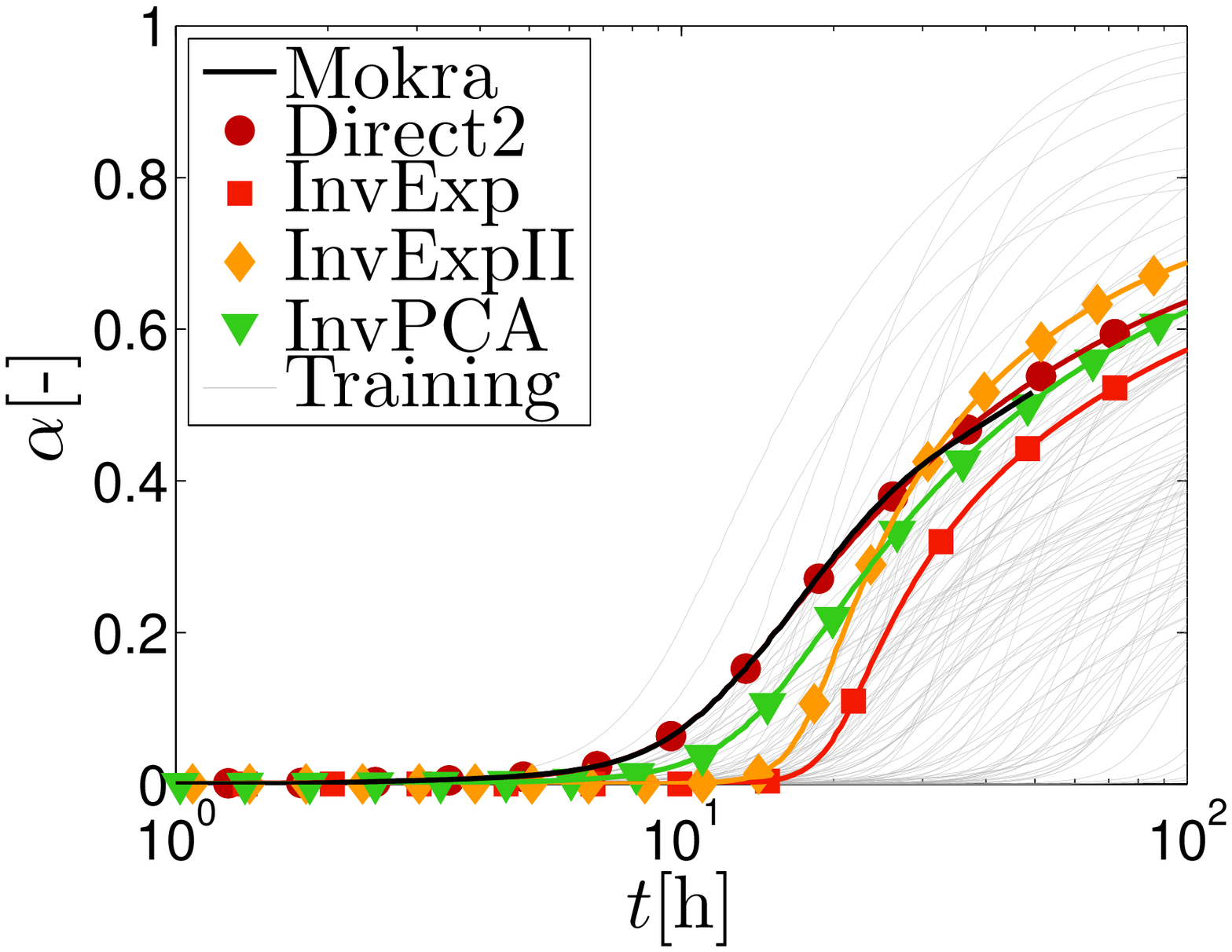}\\
\includegraphics[width=7cm]{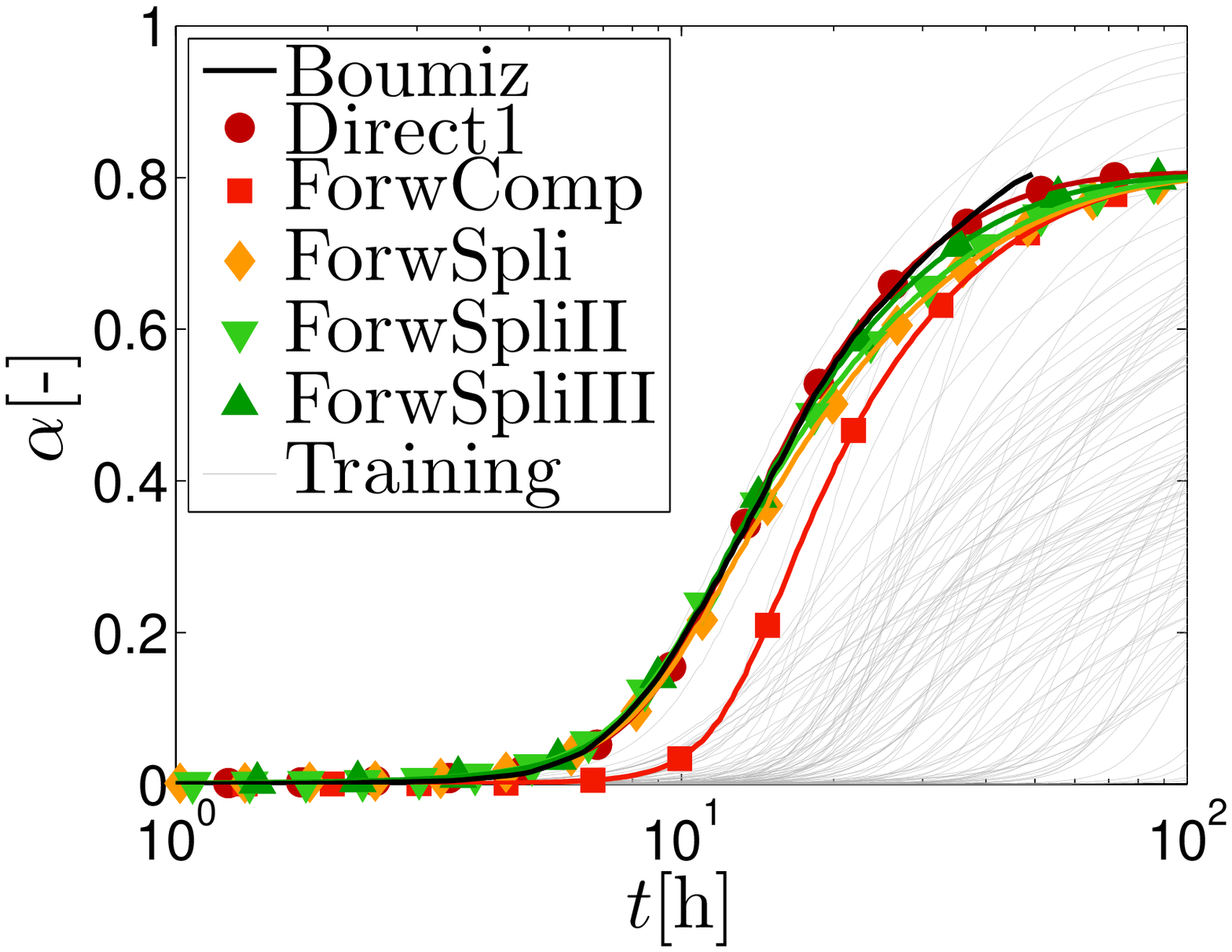} &
\includegraphics[width=7cm]{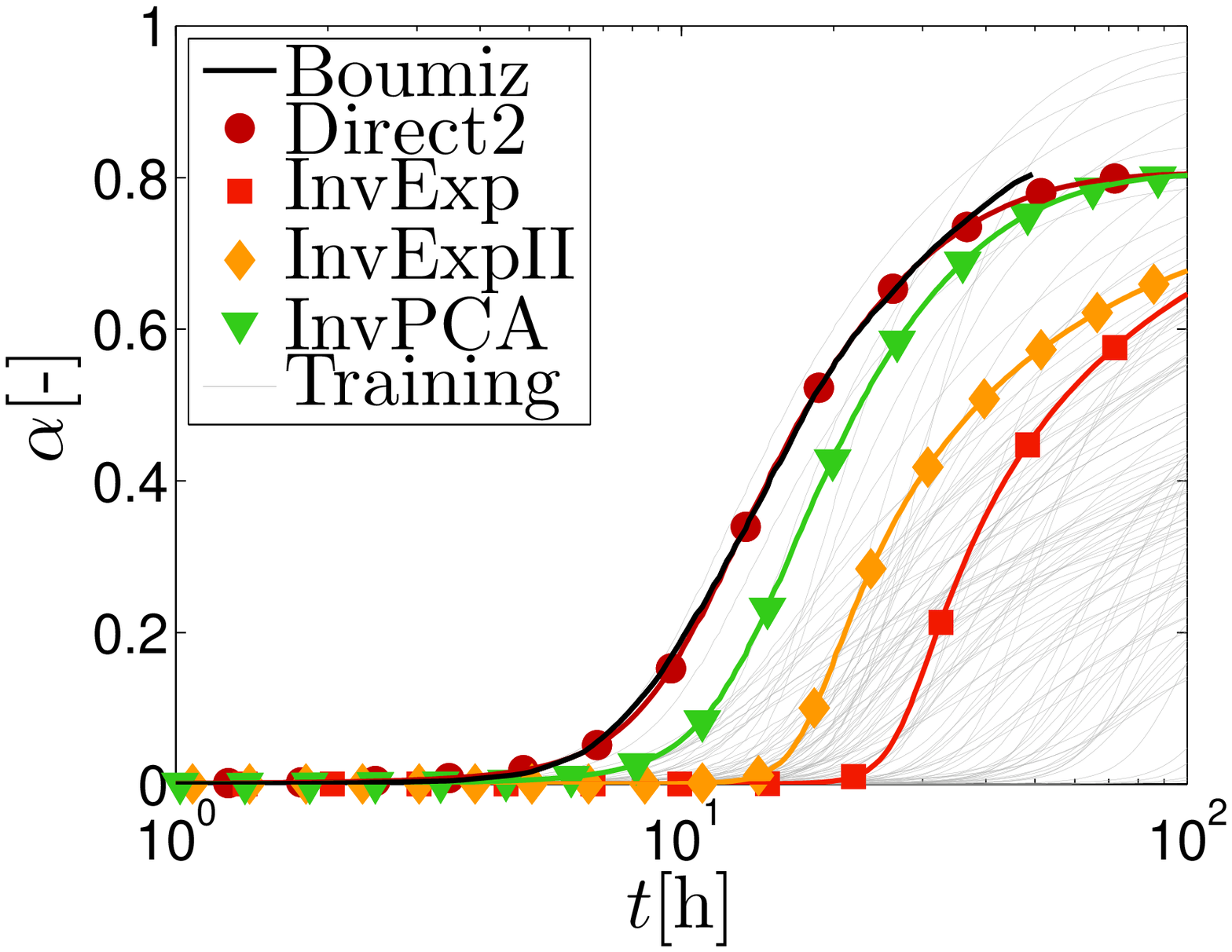}\\
\end{tabular}
\caption{Comparison of corrected experimental data ``Mokra'' and ``Boumiz'' and
  corresponding results of calibration strategies.}
\label{fig:valid1}
\end{figure}

\begin{figure}[h!]
\begin{tabular}{cc}
\includegraphics[width=7cm]{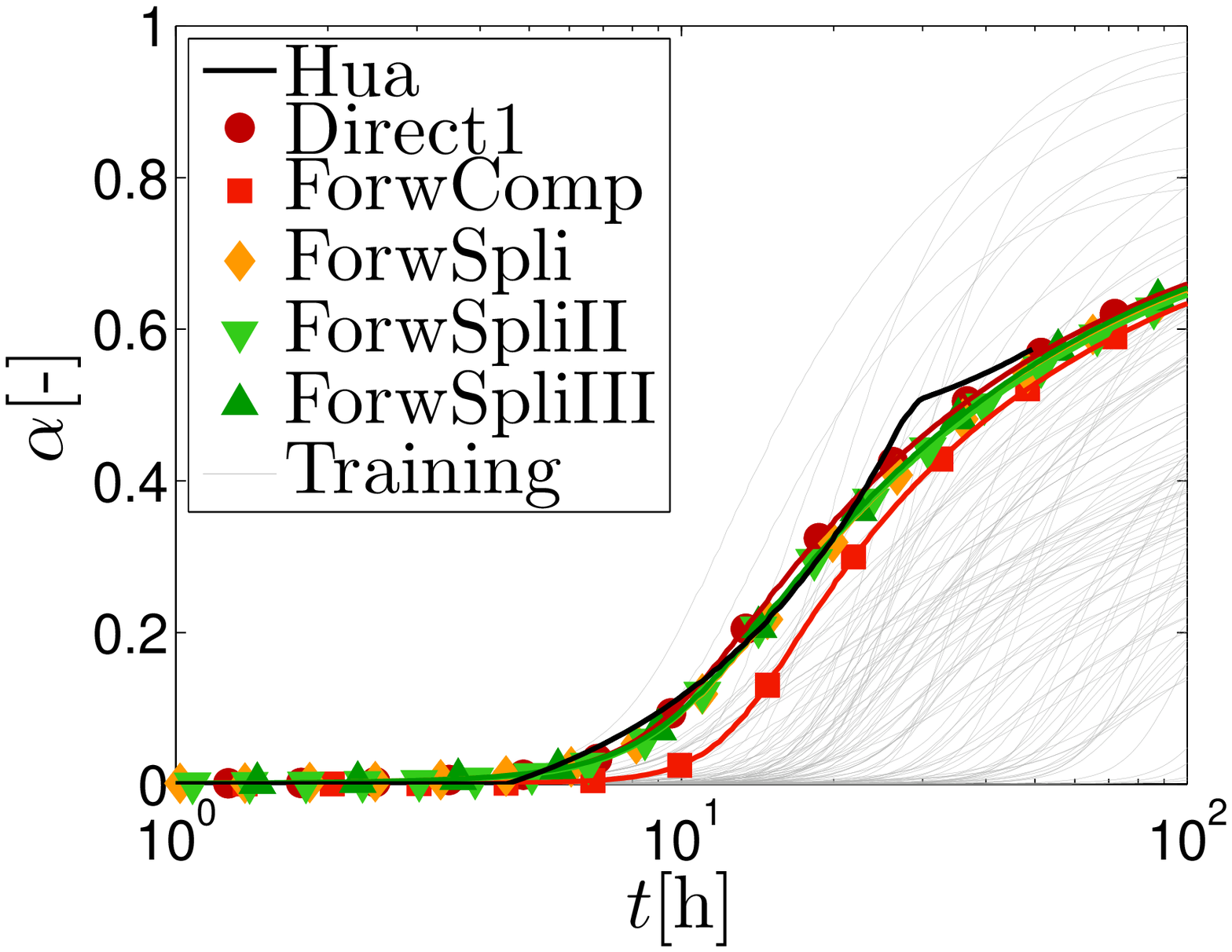} &
\includegraphics[width=7cm]{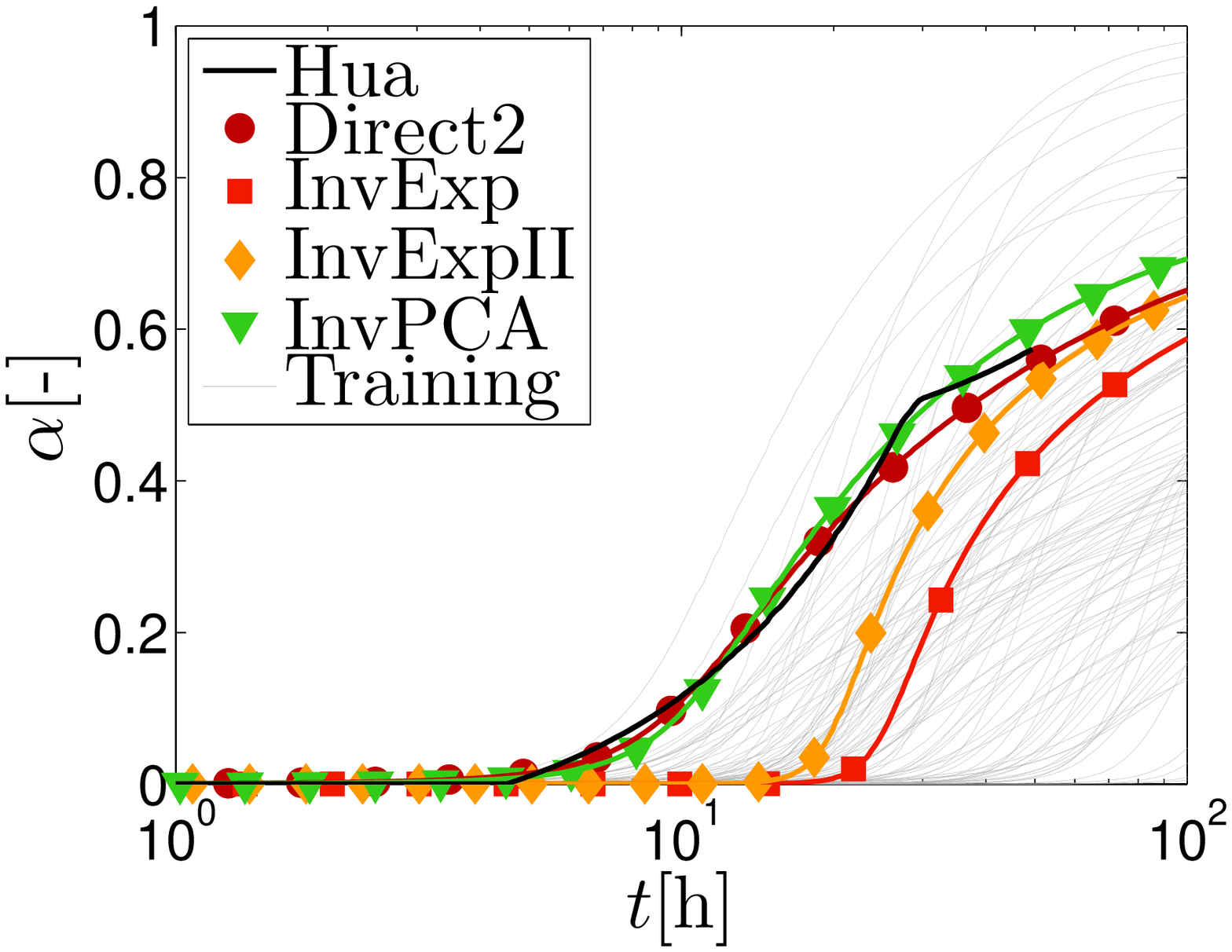}\\
\includegraphics[width=7cm]{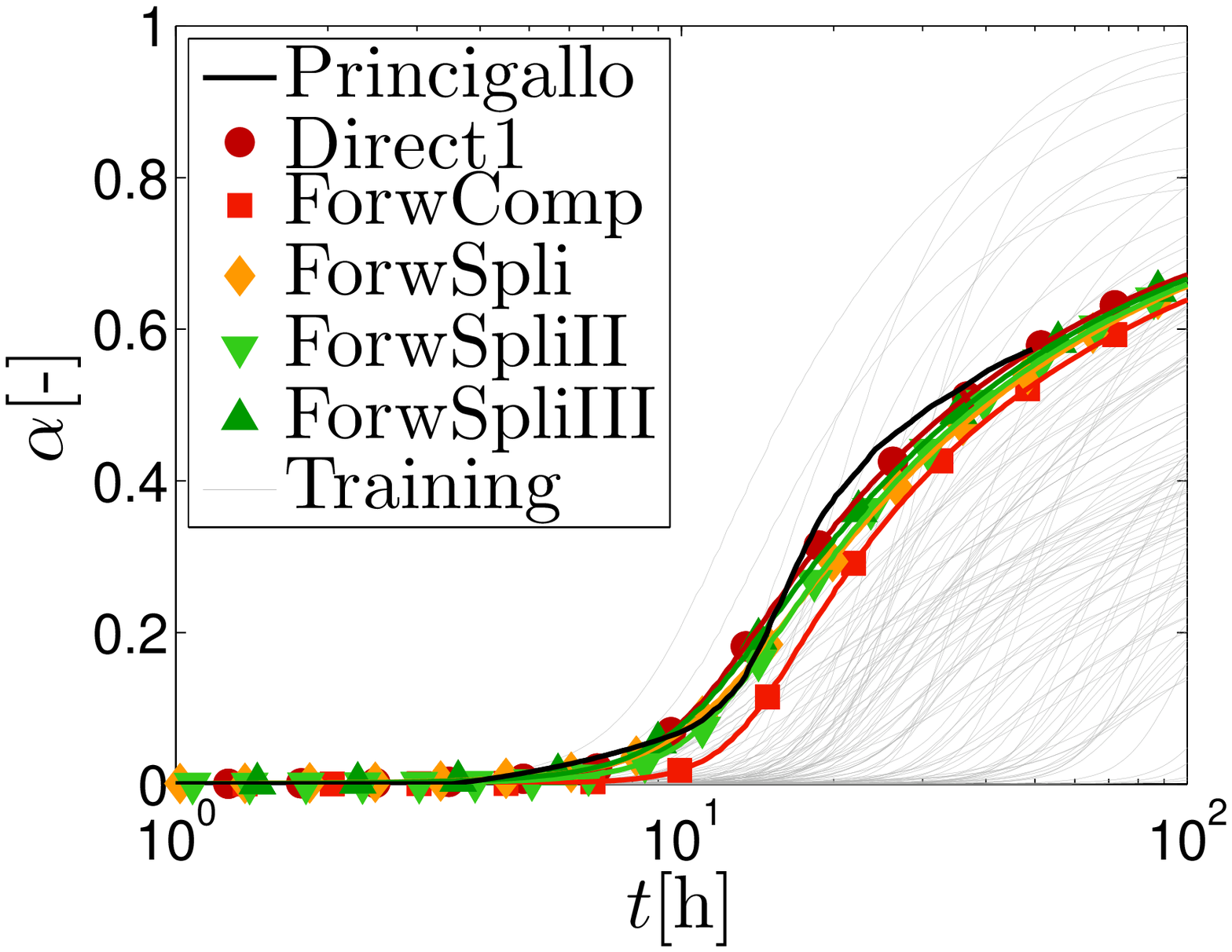} &
\includegraphics[width=7cm]{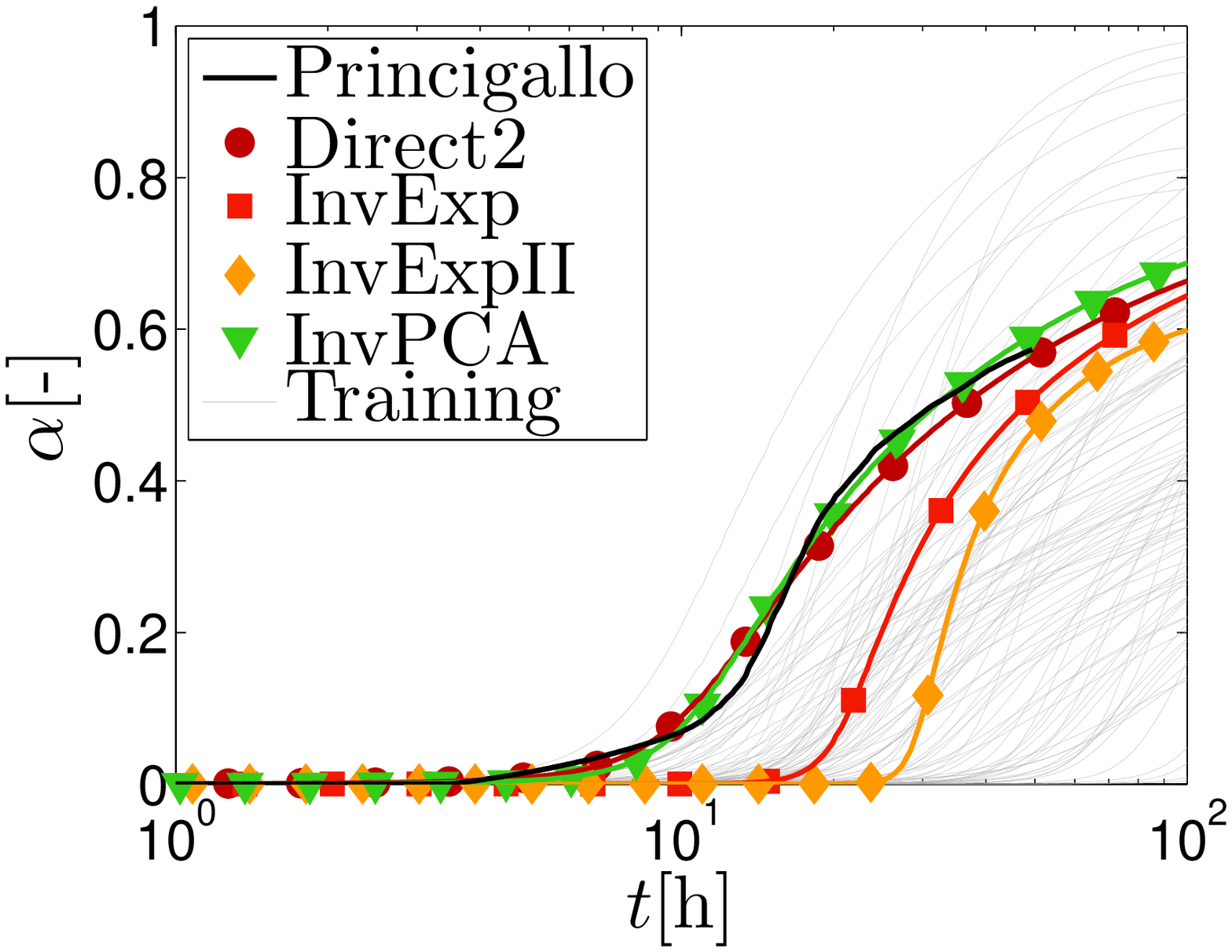}\\
\end{tabular}
\caption{Comparison of corrected experimental data ``Hua'' and
  ``Princigallo'' and corresponding results of calibration strategies.}
\label{fig:valid2}
\end{figure}

Subsequently, the identification strategies were applied to the
experimental data using the prepared ANNs. Since the ANNs are
constructed for specific time steps of the hydration degree, the
experimental curves are interpolated to the time steps required by the
particular ANNs. If necessary, the data are extrapolated beyond the
last measured time step assuming the further progress of hydration to
be constant at the last measured value. The identified parameters
together with the parameters' values obtained by the direct
optimisation are written in Tables~\ref{tab:valid1}
and~\ref{tab:valid2}. \new{Note that the parameter values highlighted
  in bold font refer to situation, where the measured data lie beyond
  the domain of training data and the ANN is forced to extrapolate.}
\begin{table}[h!]
\begin{tabular}{l|rrrr|r|rrrr|r}
\hline
 & \multicolumn{5}{l|}{``Mokra''} & \multicolumn{5}{l}{$\quad$ ``Boumiz''} \\
Method & $p_1$ & $p_2$ & $p_3$ & $p_4$ & $\widehat{\varepsilon}(\widehat{\alpha})$         & $p_1$ & $p_2$ & $p_3$ & $p_4$ & $\widehat{\varepsilon}(\widehat{\alpha})$ \\ \hline
Direct1        & 0.84 & 0.99 & 0.18 & 0.05   & 0.70      & $\quad$ 0.93 & 1.00 & 0.02 & 0.36   & 2.37 \\
Direct2        & 0.82 & 0.98 & 0.18 & 0.05   & 0.65      & 0.93 & 1.00 & 0.02 & 0.35   & 2.70 \\
\hline
Forward Complex       & 0.81 & 1.00 &  0.18 & 0.03  & 1.35      & 1.00 & 0.61 &  0.08 & 0.36  & 12.67  \\
Forward Split       & 0.82 & 1.00 &  0.19 & 0.05  & 1.15      & 0.96 & 1.00 &  0.08 & 0.35  & 5.44  \\
Forward Split II     & 0.78 & {\bf 1.01} &  0.18 & 0.05  & 0.83      & 1.00 & 1.00 &  0.08 & 0.35  & 4.11  \\
Forward Split III     & 0.80 & 1.00 &  0.19 & 0.05  & 0.91      & 0.98 & 1.00 &  0.05 & 0.35  & 3.03  \\
\hline
Error $F_1$     & 0.78 & 0.73 &  0.09 & 0.07  & 3.89      & - & - & - & - & -  \\
Error $F_2$     & 1.00 & {\bf 1.19} &  0.15 & {\bf -0.06} & 2.73       & - & - & - & - & -  \\
\hline
Inverse Expert         & {\bf 1.16} & {\bf -0.18} &  0.29 & 0.03 & 6.83      & 0.78 & {\bf -0.24} &  0.22 & 0.30 & 35.11 \\
Inverse Expert II       & {\bf 1.21} & {\bf -0.06} &  0.19 & 0.16 & 4.68      & {\bf 1.27} & {\bf -0.14} &  0.20 & 0.13  & 25.94 \\
Inverse PCA         & 0.75 & 0.83 &  0.18 & 0.06  & 1.82      & 0.78 & 0.87 &  0.02 & 0.35  & 10.82  \\
\hline
\end{tabular}
\caption{Results of identification strategies obtained for ``Mokra''
  and ``Boumiz'': identified values of model parameters and mean
  relative error in degree of hydration
  $\widehat{\varepsilon}(\widehat{\alpha})$ [\%].}
\label{tab:valid1}
\end{table}
\begin{table}[h!]
\begin{tabular}{l|rrrr|r|rrrr|r}
\hline
 & \multicolumn{5}{l|}{``Hua''} & \multicolumn{5}{l}{$\quad$ ``Princigallo''} \\
Method & $p_1$ & $p_2$ & $p_3$ & $p_4$ & $\widehat{\varepsilon}(\widehat{\alpha})$       & $p_1$ & $p_2$ & $p_3$ & $p_4$ & $\widehat{\varepsilon}(\widehat{\alpha})$  \\\hline
Direct1        & 1.00 & 0.94 & 0.20 & 0.11   & 2.24     & $\quad$ 1.00 & 0.85 & 0.19 & 0.14  & 3.46 \\
Direct2        & 0.99 & 0.96 & 0.21 & 0.11   & 2.46     & 1.00 & 0.88 & 0.21 & 0.15  & 3.27 \\
\hline
Forward Complex       & 1.00 & 0.64 &  0.22 & 0.08  & 4.10     & 1.00 & 0.58 &  0.23 & 0.14 & 6.21 \\
Forward Split       & 0.87 & 1.00 &  0.19 & 0.11  & 2.84     & 0.78 & 0.98 &  0.18 & 0.15 & 4.39 \\
Forward Split II     & 0.93 & 0.96 &  0.21 & 0.11  & 2.92     & 0.92 & 0.82 &  0.20 & 0.14 & 4.44 \\
Forward Split III     & 0.87 & {\bf 1.01} &  0.18 & 0.10  & 2.71     & 0.89 & 0.92 &  0.18 & 0.14 & 3.75 \\
\hline
Inverse Expert         & 0.94 & {\bf -0.29} &  0.26 & 0.12 & 10.64    & {\bf 1.07} & {\bf -0.16} &  0.22 & 0.15 & 9.02\\
Inverse Expert II       & {\bf 1.26} & {\bf -0.27} &  0.19 & 0.02 & 6.23     & {\bf 1.52} & {\bf -1.38} &  0.13 & {\bf -0.24} & 15.05\\
Inverse PCA         & 1.00 & 0.89 &  0.15 & 0.12  & 2.41     & {\bf 1.13} & 0.74 &  0.19 & 0.15 & 3.62 \\
\hline
\end{tabular}
\caption{Results of identification strategies obtained for ``Hua''
  and ``Princigallo'': identified values of model parameters and mean relative error
  in degree of hydration $\widehat{\varepsilon}(\widehat{\alpha})$ [\%].}
\label{tab:valid2}
\end{table}
The identified parameters were used as inputs for simulations, whose
results are compared with the experimental data in
Figures~\ref{fig:valid1} and \ref{fig:valid2}. To quantify the quality
of obtained fits, Tables~\ref{tab:valid1} and \ref{tab:valid2} contain
also the mean relative error $\widehat{\varepsilon}(\widehat{\alpha})$
[\%] computed in the same manner as in Table~\ref{tab:ident} for an
easy comparison of the verification and validation results.

The strategies based on the error function approximation are
illustrated on the parameter identification from ``Mokra'' data, which are
used to define the error functions, which are approximated by the
ANNs. The trained ANNs are then optimised by the GRADE algorithm so as to
provide the optimal set of the identified parameters.  As we presumed, the
identification results are not satisfactory despite very good results
of the ANNs' training processes, see Table~\ref{tab:training}. The
training and testing errors are small relatively to the spread of
error functions' values, which increase quickly with the distance from
the optimal solution. The strategy, however, requires high precision
of the ANN's approximation near the optimal solution, which can be
hardly achieved due the overall complex shape of the error functions.

The worst results on all the experimental curves were obtained by the inverse
strategies based on selected components of the model response used as
the ANNs' inputs. The results pointed out the high sensitivity of this
strategy to the measurement noise and to the specific choice of the inputs.
Both drawbacks are overcome by employing principal component analysis,
which allows to employ a high number of the response components and filter
the measurement noise out of the several first principal
components. The Inverse PCA strategy thus achieved significantly
better results.

The forward strategies provided generally the best results consistent with the
results of verification on the simulated data. These strategies thus
proved to be rather immune to the noise in experimental data.

\section{Conclusions}\label{sec:concl}
\label{concl}

The presented paper reviews and compares several possible applications
of artificial neural networks in calibration of numerical models.  In
particular, the feedforward layered neural network is employed in
three basic schemes to surrogate: (i) response of a model, (ii)
inverse relationship of model parameters and model response and (iii)
error function quantifying how well the model response fits the
experimental data. Their advantages and drawbacks are illustrated on
calibration of four parameters of the affinity hydration model. The
model is chosen for its nonlinearity, difference in sensitivities to
particular parameters on one hand and simplicity and very fast
numerical evaluation on the other. The later allow for the model
calibration based on the stochastic evolutionary algorithm without any
involved approximation and thus better quantification of calibration
results provided by the particular strategies. The investigated calibration
strategies are verified on $50$ simulated curves of hydration degree
and validated on four experimental ones.

\begin{table}[h!]
\centering
\begin{tabular}{lccccc}
\hline
Strategy        & $N_{\mathrm{ANN}}$   &  optimisation & new data     & errors \\
\hline
Forward Complex & $1$               &  yes  & optimisation         & middle \\
Forward Split   & $N_{\mathrm{\alpha}}$ &  yes  & optimisation         & low    \\
Error $F$       & $1$               &  yes  & traininng + optimisation & high   \\
Inverse Expert  & $N_{\mathrm{p}}$     &  no   & -             & high   \\
Inverse PCA     & $N_{\mathrm{p}}$     &  no   & -             & middle  \\
\hline
\end{tabular}
\caption{Simplified summary of calibration
  strategies. $N_{\mathrm{\alpha}}$ stands for a number of approximated
  components of model response, $N_{\mathrm{p}}$ is a number of model parameters.}
\label{tab:summary}
\end{table}

Simplified summary of the obtained results is written in
Table~\ref{tab:summary}. One of the simplest strategies from the
implementation point of view is based on an approximation of the error
function (Error~$F$), where only one neural network needs to be
trained for the prediction of the error function values. This
simplicity, however, does not hold in case of multiple experimental
measurements, where the whole identification process including the
neural network training as well as its optimisation needs to be done
all over again for any new experiment. Moreover, the presented
examples revealed that the complexity of the error function may cause
difficulties for neural network training resulting in high errors in
the identified parameters. The potential of the neural network is wasted
on approximating the whole domain, while the accurate predictions are
required only in the vicinity of the optimal values of
parameters. Hence, this strategy is more suited for surrogate models
based on radial basis function networks or kriging, which can be
trained along with the optimisation of the error function thus
allowing to improve the precision in the promising area, see
e.g. \cite{Kucerova:2009}.

An equally simple strategy is based on the approximation of the model
response, where time or space variables are included among the neural
network inputs (Forward Complex). This strategy is better suited for
layered neural networks, which is trained only once and then can be
used repeatedly for any new observations. The effort invested into the
approximation of the whole domain is thus not wasted. The application
to new data requires only one new optimisation process. The results
obtained by this strategy were not excellent, but can be considered as
satisfactory solution at a low price.

The best results were achieved by separate approximations of
particular response components, where a higher number of neural
networks is trained to approximate rather simple relationship defined
by the calibrated model (Forward Split). This procedure requires more
work on networks preparation, which is compensated by high accuracy of
the obtained results. The accuracy is proportionally increasing with
the number of approximated response components and can be thus
influenced by work invested to the surrogate construction. Moreover,
the constructed approximations can be then used again for any new
data, where only the optimisation of model parameters needs to be
repeated.

The worst results were obtained by the strategy approximating the inverse
mapping from the response components to the model parameters (Inverse
Expert). Such relationship does not have to exist and can be hardly
approximated. Moreover, if the inputs for a neural network are not
properly selected and thus highly sensitive to the measurement error, the
procedure provides unsatisfactory results. Nevertheless, using an
expert knowledge for a proper selection of inputs as presented
in~\cite{Kucerova:2014:AES}, this strategy gives good results at a
very low price, since neither training nor optimisation process, but
only a simple evaluation of the trained networks is needed for parameter
identification from new data.

The necessity of the expert knowledge and sensitivity to the measurement
error can be easily circumvented by employing principal component
analysis on the model response components (Inverse~PCA). Then only the
number of components entering as inputs in the neural network needs to
be selected. The strategy thus represents a compromise solution
providing satisfactory results at a~low price especially in the repeated
application to new observed data.

\section*{Acknowledgment}
The financial support of this work by the Czech Science Foundation
(project No. 16-11473Y) is gratefully
acknowledged. We would like also to thank V\'it \v{S}milauer (CTU in
Prague) for providing us with a code of affinity hydration model,
experimental data and helpful advices.

\bibliographystyle{elsarticle-num}
\bibliography{liter}

\appendix
\section{Configurations and results of particular neural networks}
\label{app:ann_config}

The particular choice of ANN inputs and outputs
are presented in Tables \ref{tab:training_forw} and
\ref{tab:training_inv} for forward and inverse mode strategies,
respectively.
\begin{table}[h!] \tabcolsep=2pt
\centering
\begin{tabular*}{\textwidth}{@{\extracolsep{\fill} }llrlrr}\hline
  Strategy & Inputs & $h$ & Output & $\varepsilon^\mathrm{MRP}(\mathcal{D}_\mathrm{train}) [\%]$ & $\varepsilon^\mathrm{MRP}(\mathcal{D}_\mathrm{test}) [\%]$\\
  \hline
  Forward Complex & $p_1$, $p_2$, $p_3$, $p_4$, $t_k$
& 7  & $\alpha_k$ & 2.03 & 2.67 \\
  \hline
  Forward Split & $p_1$, $p_2$, $p_3$, $p_4$ & 7  & $\alpha_{300}$  & 0.06 & 0.06\\
           & $p_1$, $p_2$, $p_3$, $p_4$ & 8  & $\alpha_{400}$  & 0.07 & 0.12\\
           & $p_1$, $p_2$, $p_3$, $p_4$ & 7  & $\alpha_{500}$  & 0.08 & 0.11\\
           & $p_1$, $p_2$, $p_3$, $p_4$ & 2  & $\alpha_{600}$  & 0.62 & 1.01\\
           & $p_1$, $p_2$, $p_3$, $p_4$ & 7  & $\alpha_{700}$  & 0.79 & 1.01\\
           & $p_1$, $p_2$, $p_3$, $p_4$ & 6  & $\alpha_{800}$  & 1.06 & 1.27\\
           & $p_1$, $p_2$, $p_3$, $p_4$ & 8  & $\alpha_{900}$  & 0.28 & 0.32\\
           & $p_1$, $p_2$, $p_3$, $p_4$ & 10 & $\alpha_{1000}$ & 0.22 & 0.27\\
           & $p_1$, $p_2$, $p_3$, $p_4$ & 9  & $\alpha_{1100}$ & 0.21 & 0.30\\
 \hline
  Forward Split II & $p_1$, $p_2$, $p_3$, $p_4$ & 7  & $\alpha_{100}$  & 0.52 & 0.88\\
             & $p_1$, $p_2$, $p_3$, $p_4$ & 4  & $\alpha_{150}$  & 0.86 & 1.39\\
             & $p_1$, $p_2$, $p_3$, $p_4$ & 4  & $\alpha_{200}$  & 0.08 & 0.11\\
             & $p_1$, $p_2$, $p_3$, $p_4$ & 8  & $\alpha_{250}$  & 0.68 & 0.80\\
             & $p_1$, $p_2$, $p_3$, $p_4$ & 4  & $\alpha_{300}$  & 0.44 & 0.60\\
             & $p_1$, $p_2$, $p_3$, $p_4$ & 5  & $\alpha_{350}$  & 0.48 & 0.95\\
             & $p_1$, $p_2$, $p_3$, $p_4$ & 4  & $\alpha_{400}$  & 0.06 & 0.07\\
             & $p_1$, $p_2$, $p_3$, $p_4$ & 6  & $\alpha_{450}$  & 0.07 & 0.10\\
             & $p_1$, $p_2$, $p_3$, $p_4$ & 6  & $\alpha_{500}$  & 0.07 & 0.13\\
             & $p_1$, $p_2$, $p_3$, $p_4$ & 9  & $\alpha_{550}$  & 0.15 & 0.22\\
             & $p_1$, $p_2$, $p_3$, $p_4$ & 6  & $\alpha_{600}$  & 1.42 & 2.04\\
             & $p_1$, $p_2$, $p_3$, $p_4$ & 5  & $\alpha_{650}$  & 0.84 & 1.19\\
             & $p_1$, $p_2$, $p_3$, $p_4$ & 6  & $\alpha_{700}$  & 0.55 & 0.73\\
             & $p_1$, $p_2$, $p_3$, $p_4$ & 8  & $\alpha_{750}$  & 0.60 & 1.18\\
             & $p_1$, $p_2$, $p_3$, $p_4$ & 7  & $\alpha_{800}$  & 0.46 & 0.62\\
             & $p_1$, $p_2$, $p_3$, $p_4$ & 9  & $\alpha_{850}$  & 0.75 & 1.09\\
             & $p_1$, $p_2$, $p_3$, $p_4$ & 7  & $\alpha_{900}$  & 0.20 & 0.23\\
             & $p_1$, $p_2$, $p_3$, $p_4$ & 13 & $\alpha_{950}$  & 0.28 & 0.43\\
             & $p_1$, $p_2$, $p_3$, $p_4$ & 9  & $\alpha_{1000}$ & 0.73 & 1.16\\
             & $p_1$, $p_2$, $p_3$, $p_4$ & 6  & $\alpha_{1050}$ & 0.10 & 0.17\\
             & $p_1$, $p_2$, $p_3$, $p_4$ & 9  & $\alpha_{1100}$ & 0.10 & 0.19\\
             & $p_1$, $p_2$, $p_3$, $p_4$ & 7  & $\alpha_{1150}$ & 0.08 & 0.14\\
 \hline
 Error $F_1$  & $p_1$, $p_2$, $p_3$, $p_4$ & 10 & $F_1$ for Mokra & 0.54 & 0.74 \\
             & $p_1$, $p_2$, $p_3$, $p_4$ & 10 & $F_1$ for shifted Mokra & 0.40 & 0.57\\
 \hline
 Error $F_2$  & $p_1$, $p_2$, $p_3$, $p_4$ &  9 & $F_2$ for Mokra & 0.78 & 0.96\\
             & $p_1$, $p_2$, $p_3$, $p_4$ &  9 & $F_2$ for shifted Mokra & 1.36 & 1.56\\
    \hline
\end{tabular*}
\caption{Architecture of particular ANNs constructed in forward strategies and their errors on training and testing data.}
\label{tab:training_forw}
\end{table}

\begin{table}[h!] \tabcolsep=2pt
\centering
\begin{tabular*}{\textwidth}{@{\extracolsep{\fill} }llrlrr}\hline
  Strategy & Inputs & $h$ & Output & $\varepsilon^\mathrm{MRP}(\mathcal{D}_\mathrm{train}) [\%]$ & $\varepsilon^\mathrm{MRP}(\mathcal{D}_\mathrm{test}) [\%]$\\
\hline
Forward Split III
& $p_1$, $p_2$, $p_3$, $p_4$ & 7 & $\alpha_{100}$  & 0.07 & 0.08\\
& $p_1$, $p_2$, $p_3$, $p_4$ & 4 & $\alpha_{130}$  & 0.56 & 0.45\\
& $p_1$, $p_2$, $p_3$, $p_4$ & 4 & $\alpha_{150}$  & 1.03 & 0.95\\
& $p_1$, $p_2$, $p_3$, $p_4$ & 8 & $\alpha_{170}$  & 0.08 & 0.06\\
& $p_1$, $p_2$, $p_3$, $p_4$ & 4 & $\alpha_{200}$  & 0.76 & 0.74\\
& $p_1$, $p_2$, $p_3$, $p_4$ & 5 & $\alpha_{230}$  & 0.44 & 0.41\\
& $p_1$, $p_2$, $p_3$, $p_4$ & 4 & $\alpha_{250}$  & 0.49 & 0.45\\
& $p_1$, $p_2$, $p_3$, $p_4$ & 6 & $\alpha_{270}$  & 0.07 & 0.06\\
& $p_1$, $p_2$, $p_3$, $p_4$ & 6 & $\alpha_{300}$  & 0.07 & 0.05\\
& $p_1$, $p_2$, $p_3$, $p_4$ & 9 & $\alpha_{330}$  & 0.07 & 0.11\\
& $p_1$, $p_2$, $p_3$, $p_4$ & 6 & $\alpha_{350}$  & 0.16 & 0.37\\
& $p_1$, $p_2$, $p_3$, $p_4$ & 5 & $\alpha_{370}$  & 1.47 & 1.90\\
& $p_1$, $p_2$, $p_3$, $p_4$ & 6 & $\alpha_{400}$  & 0.84 & 1.06\\
& $p_1$, $p_2$, $p_3$, $p_4$ & 8 & $\alpha_{430}$  & 0.59 & 0.96\\
& $p_1$, $p_2$, $p_3$, $p_4$ & 7 & $\alpha_{450}$  & 0.71 & 0.92\\
& $p_1$, $p_2$, $p_3$, $p_4$ & 9 & $\alpha_{470}$  & 0.54 & 0.55\\
& $p_1$, $p_2$, $p_3$, $p_4$ & 7 & $\alpha_{500}$  & 0.89 & 0.98\\
& $p_1$, $p_2$, $p_3$, $p_4$ & 13 & $\alpha_{530}$  & 0.23 & 0.40\\
& $p_1$, $p_2$, $p_3$, $p_4$ & 9 & $\alpha_{550}$  & 0.30 & 0.44\\
& $p_1$, $p_2$, $p_3$, $p_4$ & 6 & $\alpha_{570}$  & 0.73 & 0.63\\
& $p_1$, $p_2$, $p_3$, $p_4$ & 9 & $\alpha_{600}$  & 0.12 & 0.20\\
& $p_1$, $p_2$, $p_3$, $p_4$ & 7 & $\alpha_{630}$  & 0.11 & 0.18\\
& $p_1$, $p_2$, $p_3$, $p_4$ & 7 & $\alpha_{650}$  & 0.07 & 0.08\\
& $p_1$, $p_2$, $p_3$, $p_4$ & 4 & $\alpha_{670}$  & 0.55 & 0.49\\
& $p_1$, $p_2$, $p_3$, $p_4$ & 6 & $\alpha_{700}$  & 0.07 & 0.09\\
& $p_1$, $p_2$, $p_3$, $p_4$ & 8 & $\alpha_{730}$  & 0.06 & 0.06\\
& $p_1$, $p_2$, $p_3$, $p_4$ & 9 & $\alpha_{750}$  & 0.07 & 0.06\\
& $p_1$, $p_2$, $p_3$, $p_4$ & 8 & $\alpha_{770}$  & 0.03 & 0.03\\
& $p_1$, $p_2$, $p_3$, $p_4$ & 8 & $\alpha_{800}$  & 0.05 & 0.04\\
& $p_1$, $p_2$, $p_3$, $p_4$ & 5 & $\alpha_{830}$  & 0.09 & 0.10\\
& $p_1$, $p_2$, $p_3$, $p_4$ & 5 & $\alpha_{850}$  & 0.77 & 0.42\\
& $p_1$, $p_2$, $p_3$, $p_4$ & 3 & $\alpha_{870}$  & 0.23 & 0.27\\
& $p_1$, $p_2$, $p_3$, $p_4$ & 6 & $\alpha_{900}$  & 1.06 & 0.99\\
& $p_1$, $p_2$, $p_3$, $p_4$ & 7 & $\alpha_{930}$  & 1.50 & 1.88\\
& $p_1$, $p_2$, $p_3$, $p_4$ & 8 & $\alpha_{950}$  & 0.37 & 0.49\\
& $p_1$, $p_2$, $p_3$, $p_4$ & 7 & $\alpha_{970}$  & 1.38 & 1.98\\
& $p_1$, $p_2$, $p_3$, $p_4$ & 7 & $\alpha_{1000}$  & 0.93 & 1.05\\
& $p_1$, $p_2$, $p_3$, $p_4$ & 8 & $\alpha_{1030}$  & 0.26 & 0.35\\
& $p_1$, $p_2$, $p_3$, $p_4$ & 7 & $\alpha_{1050}$  & 0.83 & 0.87\\
& $p_1$, $p_2$, $p_3$, $p_4$ & 6 & $\alpha_{1070}$  & 1.12 & 1.04\\
& $p_1$, $p_2$, $p_3$, $p_4$ & 8 & $\alpha_{1100}$  & 0.31 & 0.36\\
& $p_1$, $p_2$, $p_3$, $p_4$ & 11 & $\alpha_{1130}$  & 0.13 & 0.20\\
& $p_1$, $p_2$, $p_3$, $p_4$ & 7 & $\alpha_{1150}$  & 0.14 & 0.20\\
    \hline
\end{tabular*}
\caption{Architecture of particular ANNs constructed in forward strategies and their errors on training and testing data.}
\label{tab:training_forw2}
\end{table}

\begin{table}[h!] \tabcolsep=2pt
\centering
\begin{tabular*}{\textwidth}{@{\extracolsep{\fill}}llrlrr}\hline
  Strategy & Inputs & $h$ & Output & $\varepsilon^\mathrm{MRP}(\mathcal{D}_\mathrm{train}) [\%]$ & $\varepsilon^\mathrm{MRP}(\mathcal{D}_\mathrm{test}) [\%]$\\
  \hline
   Inverse Expert & $9$ values:  $\alpha_{300}, \alpha_{400}, \dots, \alpha_{1100}$ & 5 & $p_1$ & 5.74 & 6.43\\
    & $9$ values: $\alpha_{300}, \alpha_{400}, \dots, \alpha_{1100}$ &        7 & $p_2$ & 5.15 & 6.21\\
    & $9$ values: $\alpha_{300}, \alpha_{400}, \dots, \alpha_{1100}$ &        8 & $p_3$ & 1.99 & 2.16\\
    & $9$ values: $\alpha_{300}, \alpha_{400}, \dots, \alpha_{1100}$ &        5 & $p_4$ & 1.14 & 1.31\\
    \hline
    Inverse Expert II & $10$ values: $\alpha_{200}, \alpha_{300}, \dots, \alpha_{1100}$ & 5 & $p_1$ & 5.79 & 6.23\\
    & $10$ values: $\alpha_{200}, \alpha_{300}, \dots, \alpha_{1100}$ &          4 & $p_2$ & 5.60 & 6.52\\
    & $10$ values: $\alpha_{200}, \alpha_{300}, \dots, \alpha_{1100}$ &          6 & $p_3$ & 2.60 & 3.18\\
    & $10$ values: $\alpha_{200}, \alpha_{300}, \dots, \alpha_{1100}$ &          5 & $p_4$ & 1.38 & 1.36\\
    \hline
    Inverse PCA &  $9$ values: $\bar{\alpha}_1, \bar{\alpha}_2, \dots, \bar{\alpha}_9$ & 6 & $p_1$ & 3.86 & 5.10\\
    & $9$ values: $\bar{\alpha}_1, \bar{\alpha}_2, \dots, \bar{\alpha}_9$         & 4 & $p_2$ & 10.50 & 16.73\\
    & $9$ values: $\bar{\alpha}_1, \bar{\alpha}_2, \dots, \bar{\alpha}_9$         & 8 & $p_3$ & 1.25 & 1.89\\
    & $9$ values: $\bar{\alpha}_1, \bar{\alpha}_2, \dots, \bar{\alpha}_9$         & 8 & $p_4$ & 0.28 & 0.33\\
    \hline
\end{tabular*}
\caption{Architecture of particular ANNs in inverse strategies and their errors on training and testing data.}
\label{tab:training_inv}
\end{table}

\end{document}